# Detecting Potentially Harmful and Protective Suicide-related Content on Twitter: A Machine Learning Approach

**Authors**: Hannah Metzler[1,2,3,4,5], Hubert Baginski[3,6], Thomas Niederkrotenthaler[2], David Garcia[1,3,4]

**Affiliations:**

1. Section for Science of Complex Systems, Center for Medical Statistics, Informatics and Intelligent Systems, Medical University of Vienna, Austria
2. Unit Suicide Research & Mental Health Promotion, Department of Social and Preventive Medicine, Center for Public Health, Medical University of Vienna, Austria
3. Complexity Science Hub Vienna, Austria
4. Institute of Interactive Systems and Data Science, Department of Computer Science and Biomedical Engineering, Graz University of Technology, Graz, Austria
5. Institute of Globally Distributed Open Research and Education, Austria
6. Institute of Information Systems Engineering, Vienna University of Technology, Vienna, Austria

# Abstract

### Background

Research has repeatedly shown that exposure to suicide-related news media content is associated with suicide rates, with some content characteristics likely having harmful and others potentially protective effects. Although good evidence exists for a few selected characteristics, systematic and large-scale investigations are missing. Moreover, the growing importance of social media, particularly among young adults, calls for studies on the effects of content posted on these platforms.

### Objective

This study applies natural language processing and machine learning methods to classify large quantities of social media data according to characteristics identified as potentially harmful or beneficial in media effects research on suicide and prevention.

### Methods

We manually labeled 3202 English tweets using a novel annotation scheme that classifies suicide-related tweets into 12 categories. Based on these categories, we trained a benchmark of machine learning models for a multi-class and a binary classification task. As models, we included a majority classifier, an approach based on word frequency (TF-IDF with a linear SVM) and two state-of-the-art deep learning models (BERT, XLNet). The first task classified postings into six main content categories, which are particularly relevant for suicide prevention based on previous evidence. These included personal stories of either (1) suicidal ideation and attempts or (2) coping and recovery, calls for action intending to spread either (3) problem awareness or (4) prevention-related information, (5) reporting of suicide cases, and (6) other tweets irrelevant to these five categories. The second classification task was binary, and separated postings in the 11 categories that refer to actual suicide, from postings in the off-topic category, which use suicide-related terms in another meaning or context.

### Results

In both tasks, the performance of the two deep learning models was very similar and better than the majority or the word frequency classifier. BERT and XLNet reached accuracy scores above 73% on average across the six main categories in the test set, and F1-scores in between 0.69 and 0.85 for all but the suicidal ideation and attempts category (F1 = 0.55). In the binary classification task, they correctly labeled around 88% of tweets as about suicide vs. off-topic, with BERT achieving F1-scores of 0.93 and 0.74, respectively. These classification performances were similar to human performance in most cases, and are comparable to the state-of-the-art on similar tasks.

### Conclusions

The achieved performance scores highlight machine learning as a useful tool for media effects research on suicide. The clear advantage of BERT and XLNet hints that there is crucial information about meaning in the context of words beyond mere word frequencies in tweets about suicide. By making data labeling more efficient, this work has enabled large-scale investigations on harmful and protective associations of social media content with suicide rates and help-seeking behavior.

## Introduction

Suicide is a major public health problem worldwide, accounting for 1.4% of all deaths, equaling almost 800.000 in 2017, and many more suicide attempts [1]. Research shows that exposure to suicide-related news media content can influence suicidal behavior in vulnerable individuals, in both a harmful and beneficial direction. If suicide cases increase or decrease after exposure to suicide-related news seems to depend on specific elements of media content and language. As a recent-meta-analysis of media effects research on suicide shows, most solid evidence exists for increases of suicides after exposure to news about celebrity deaths by suicide [2]. This imitation of suicidal behavior is commonly referred to as the Werther effect [3]. In contrast, exposure to other types of content may have a protective effect, with the strongest evidence existing for stories of hope, recovery and coping [4–6]. Yet, broader prevention texts (i.e., texts focused on prevention that were not personal stories of recovery) have also been found to be associated with protective effects in some studies [7,8]. The association of positive messaging on suicide prevention with later decreases in suicide rates has been labeled the Papageno effect [4].

Studies investigating how exposure to media content is associated with suicidal behavior have so far mainly focused on traditional news outlets, such as print and online newspapers, radio and TV broadcasts. Investigations of associations of social media content with suicides remain extremely scarce [9–13]. The majority of previous research on social media has focused on detecting suicidal ideation in users' postings with the purpose to identify individuals at risk, but very little research has been conducted to analyze media effects. The applied methods for identifying such individuals include machine learning as well as word dictionaries, word frequencies, topic models, and social network analysis [e.g., 17–21] [for more, see reviews by 22–26]. A small set of studies has started developing machine learning classifiers for content other than suicidal ideation, despite evidence from research on traditional media that other content types can affect suicidal behavior [e.g., 2,4].

One machine learning study categorized tweets according to expressed emotions [9], while two further studies [16,27] classified typically occurring content types, including celebrity suicide reports, suicidal intent, awareness campaigns, prevention information, condolences and flippant remarks. Although these two studies include several different prevention-relevant content types, they both use the same and relatively small data set, which is limited to tweets containing celebrity names, or suicidal intent. Furthermore, all of these machine learning studies have used word frequency statistics as predefined features for model training, which cannot capture differences in the meaning of words across different contexts. This study addresses several gaps in the existing literature on media effects on suicidal behavior. The first is the lack of research on suicide-related social media content other than suicidal ideation. Suicide is a leading cause of deaths among young adults [1], who predominantly get their news on such platforms [13,14]. This highlights the urgency of systematic research on social media effects. Additionally, social media postings often feature other content types than traditional news outlets, on which research is required. This includes diary-like postings in which people describe their personal experiences, or postings addressed in their social network with the intention to prevent suicides. In the present work, we investigate Twitter data, and create a detailed annotation scheme for the types of suicide-related tweets that are potentially relevant to prevention efforts.

Second, regarding prevention-related media content, there is currently a discussion in the literature on whether content that highlights prevalence data to increase problem awareness has a protective effect, or may even be detrimental [28]. By highlighting the prevalence of suicide and risk factors like mental health or abuse without mentioning solutions to the issue, attempts to spread awareness may normalize suicidal behavior and trigger harmful effects [29]. In the present work, we address the lack of studies differentiating between prevention messages focusing on prevalence vs. prevention opportunities [28], and distinguish between awareness and prevention-focused calls for action on Twitter.

Third, the samples used in previous studies on suicide-related social media content are limited either in size [e.g., 10,11] or by a set of search terms used to collect tweets [e.g., 12,15. Sample sizes are usually small since all content needs to be annotated manually. Search terms either narrowly focus on events such as the suicide of specific celebrities [e.g., 12], or broadly include all texts containing the word suicide. Thereby, the effects of different content types may cancel each other out [2]. Thus, to investigate potentially harmful and protective effects in a systematic way, a large-scale and simultaneously fine-grained approach is necessary.

We address these challenges by first developing a comprehensive annotation scheme that systematically organizes tweets about suicide into categories most likely to beneficially or harmfully affect suicide and help seeking behavior based on available evidence [e.g., 2,4,6,30]. Second, we compare different natural language processing and machine learning methods to automatically detect and classify particularly important categories in large quantities of social media data. Extending previous work on different prevention-related social media content types [9,16,27], we include not only word-frequency-based models, but also two deep learning models that can capture content-dependent meanings of words [19]. We train all models in two tasks, a multi-class classification problem with six main content categories, and a binary classification problem of tweets about actual suicide vs. off-topic tweets, which use the word suicide in another meaning or context.

The six assessed main categories include five content types that are particularly relevant for suicide prevention based on previous research.. As described above, strongest evidence exists for suicide case reports having harmful effects, and for personal stories of hope and coping having protective effects. Another type of personal stories in Twitter postings mention suicidal ideation and attempts, without any hint at coping or recovery. Preliminary evidence suggests such postings might have a protective effect [11]. Some evidence also suggests protective effects for general prevention messages [7,8]. We distinguish between general prevention messages calling for action by either spreading prevention-related information or solution-oriented attitudes, from those spreading problem awareness only. Finally, we include an irrelevant category to identify tweets outside the other five possible categories described above.

The objective of our study was to enable large scale studies on the association of tweet content with suicidal and help-seeking behavior. More specifically, we aimed at providing volume estimates for the different prevention-relevant tweet categories for follow-up studies on the associations of these estimates with the number of suicide cases and helpline calls.

# Methods

## Data Set for Training Machine Learning Models

Given that this study is part of a project about media messaging for suicide prevention in the United States of America (US), all data sets of this study include English tweets of users located in the US. We retrieved tweet IDs via the data reseller Crimson Hexagon (now known as Brandwatch) previously used for suicide research [9,12], and then downloaded the full text of these tweets via the Twitter API. Crimson Hexagon provides access to the entire history of Twitter data, and includes reliable language and location filters. The location algorithm manages to match 90% of all posts in a country to a location using a combination of geo-coordinates, location information from user profiles, and users' time zones and languages [1].

Using the keywords and exclusion terms below, we created a pool of unique tweets without duplicates or retweets, based on which we prepared a labeled set of tweets for training machine learning models.

We retrieved tweets posted in between 1 January 2013 and 31 May 2020 (see note on dates in Appendix 1), that contained at least one of the suicide-related search terms taken from a previous study [11]The search terms were: suicide, suicidal, killed himself, killed herself, kill himself, kill herself, hung himself, hung herself, took his life, took her life, take his life, take her life, end his own life, end her own life, ended his own life, ended her own life, end his life, end her life, ended his life, ended her life, ends his life, ends her life.

The exclusion terms were identified by inspecting word frequency plots for common terms that may indicate that tweets used the term suicide for describing something other than someone ending their life, or terms that indicated tweets about suicide bombing. We then verified if these terms were actual mismatches by reading example tweets containing these terms. In this way, tweets with the most common usages of the term suicide in contexts that do not refer to actual suicide could be excluded. The final list of exclusion terms was: suicide squad (a movie), suicidechrist, SuicideGirl* (a website featuring pin-up photography of models), SuicideBoy* (male models), suicideleopard (a frequently mentioned Twitter user), suicidexjockey* (a Twitter user), suicidal grind (a music album), Epstein (excessive speculations about whether the death of Jeffrey Epstein was or wasn't a suicide), political suicide (tweets using suicide as a metaphor for political failure), Trump, clinton*, Hillary, Biden, sanders (also mostly about political suicide).

To avoid over-learning from multiple identical tweets, we made sure the labeled data used for Machine Learning would not include any tweet duplicates. We excluded retweets (tweets categorized as retweets by Crimson Hexagon given the metadata of tweets, as well as tweets containing manual labels for retweets (RT), or slightly modified tweets (MT). . We assembled a labeled data set of 3202 tweets by iteratively selecting tweets from the larger pool of tweets as described in the next section. In the following, we refer to these 3202 tweets as the *total labeled data set.* While part of this data set was put together using keywords and model predictions (see below), a second sub-sample of 1000 tweets was selected randomly. We refer to these 1000 tweets as the *randomly selected labeled data set.*

In the course of the study, we put together two other data sets, the first to compare model and human inter-rater reliability and the second for a face validity check and a follow-up study [31]. Both of them are described in detail in the section Evaluating Reliability and Face Validity of Model Predictions for BERT.

## Creating the Annotation Scheme and Labeled Data Set

Creating the annotation scheme and the labeled data set was an iterative human-in-the-loop process building on preliminary classifiers and annotations. We started with five broad categories which appeared most relevant given previous research on traditional media (see Introduction). We then added additional categories when tweets did not fit into the existing categories, but might nevertheless be associated with suicides. Given that the tweets of interest are relatively rare compared to irrelevant tweets, we used the following step wise procedure to identify examples

- We manually selected approximately 100 tweets for each of the 5 main categories (550 tweets in total): *suicide cases, coping stories, awareness, prevention, and irrelevant tweets*. To gather this first set of tweets, we searched the data set for typical examples, both randomly and with keywords that might indicate each particular category. We iteratively expanded the list of keywords by inspecting the most frequent terms in the resulting tweets in a systematic way [32]. The full list of keywords is in the Supplementary Information (SI); examples are: *committed* or *found dead* for suicide cases*; recover** or *hope* for coping stories; *lifeline* or *prevention* for prevention; *awareness, please retweet* or *please copy* for awareness.

- We used a preliminary machine learning model to make predictions based on this first training data set of 550 tweets, to identify potential examples for each category. Next, two authors with domain expertise (T.N., H.M.) continued annotating 100 tweets from each of 5 predicted categories (484 after removing duplicates and missing labels from one coder). Inter-rater reliability for these 500 tweets was a Cohen Kappa of 0.75. Based on careful inspection of all disagreements, we refined the definitions for all categories, and adjusted the labels of all previously annotated tweets accordingly. Annotating these tweets, we further noticed a novel type of message not described in research on traditional news reporting, namely purely negative descriptions of suicidal experiences without any hint at coping, hope or recovery. We updated the annotation scheme to include this new category *Suicidal ideation & attempts*, resulting in 6 main categories in total. The total training set of tweets now included n=1034 tweets.
- At this stage of the labeling process, we found that two dimensions were generally helpful to differentiate between categories: the message type, (.e.g, a personal story, a news story, a call to action, etc.,) and the underlying perspective about suicide, (i.e., if the tweet applies a problem- and suffering- or solution- and coping-centered perspective). For each message type, we noticed that some tweets implicitly or explicitly frame suicide only as a problem or from an exclusively negative/suffering perspective (categories: suicidal ideation and attempts, suicide cases, awareness), while other tweets imply that coping is possible, or suggest ways of dealing with the problem (categories: coping stories, prevention).
- Repeating step 2, we trained our best preliminary model to make new predictions for the six categories based on all labeled tweets. Each coder. annotated a different set of tweets for each predicted label until we reached a minimum number of 200 training examples for the smallest categories (suicidal and coping stories). This resulted in n=2206 tweets in total.
- In order to mitigate bias from the search terms we used to assemble our initial training set, and to estimate the distribution of tweets across categories on Twitter, H.M. labeled a random sample of 996 tweets (initially 1000, 4 were not labeled due to a displaying error in the used spreadsheet). These were added to the training set, resulting in a total sample of 3202 tweets.
- After having reviewed the entire training set, we finally refined the categories to allow for the following distinctions: For stories about coping and suicidal experiences, we differentiated the perspective from which an experience was described (1st or 3rd person), which experience was described (the one of a concerned or a bereaved individual), and whether a tweet was shared by news media or individual users. For reporting of cases, we distinguished tweets about individuals who had actually died by suicide, from tweets about someone saving the life of an individual who was about to take his or her life. Finally, we organized the tweet categories according to two dimensions further described in the next section.

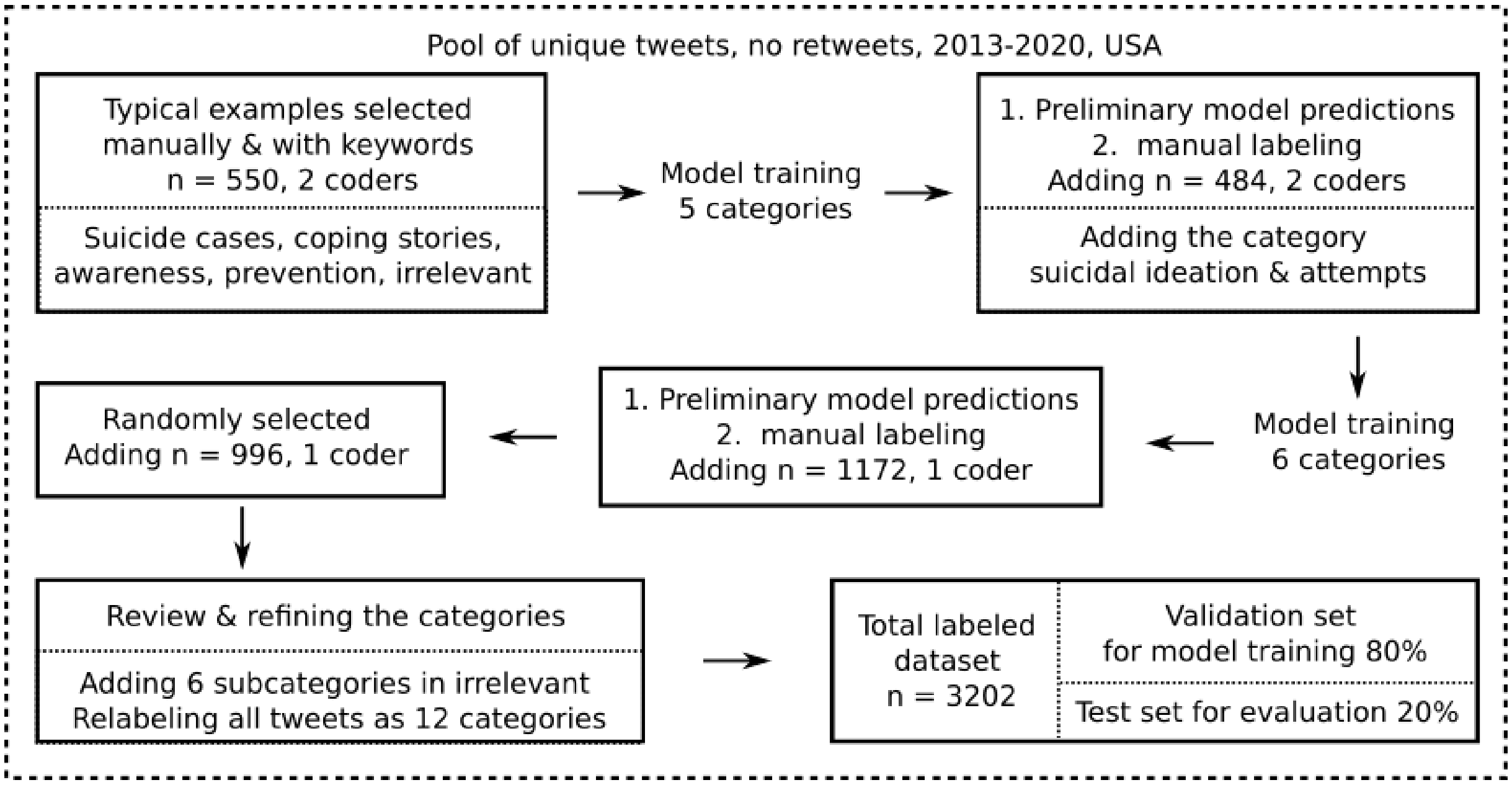

**Figure 1. Creating the labeled data set and annotation scheme.** Each box describes how tweets were selected from the large pool of available tweets, how many tweets were added to the training data set in each step (after removing duplicates), and how many coders labeled each tweet. When we used preliminary model predictions to identify potential candidates for each category, we deleted the model labels before manual coding. After rounds with 2 coders, we checked inter-rater reliability, and adapted the annotation scheme until all disagreements were clarified, and relabeled the respective sample.

## Annotation Scheme

The annotation scheme divides tweets into 12 different categories, including ten categories of interest and two irrelevant categories. Each category can be described in terms of two dimensions: the message type, .e.g, a personal story, a news story, a call to action, etc., and the underlying perspective about suicide, i.e., if the tweet applies a problem- or solution-centered perspective. The perspective distinguishes messages which implicitly or explicitly frame suicide only as a problem or from an exclusively negative/suffering perspective, from messages which imply that coping is possible, or suggest ways of dealing with the problem. The organization of tweet categories along these two dimensions is depicted in Table 1. Detailed instructions for annotating tweets are described in Appendix 2. These include prioritization rules for how to deal with ambiguous tweets that might fit into more than one category.

**Table 1. Annotation scheme of content categories organized along two dimensions: message type and underlying perspective about suicide.**

| | Underlying perspective | |
|---|---|---|
| **Message type** | **Problem & Suffering** | **Solution & Coping** |
| **Personal experiences** 1st or 3rd person | Suicidal ideation & attempts[a] | Coping (Papageno)[a] |
| **News about experiences & behavior** | News suicidal ideation & attempts | News coping |
| **Experience of bereaved** | Bereaved negative | Bereaved coping |

| | | | |
|---|---|---|---|
| **Case reports** | | Suicide cases (Werther)[a] | Lives saved |
| **Calls for action** | | Awareness[a] | Prevention[a] |
| **Irrelevant**[a] | **Suicide other** | Murder-suicides, history, fiction, not being suicidal, opinions... | |
| | **Off-topic**[b] | Bombings, euthanasia, jokes, metaphors, band/song names... | |

[a] The 6 main categories classified in machine learning Task 1
[b] *Task 2 distinguished the off-topic category from all other categories* (see section Classification Tasks).

**Description of content categories**

For each message type (except for irrelevant messages), there is a category for more problem- or suffering-focused tweets, and for more solution- or coping-focused tweets:

1. Personal stories describing the experience of an affected individual either in first or third person perspective:
   a. Suicidal ideation & attempts: Personal stories about an individuals negative experience with suicidal thoughts, related suffering (e.g., depression), suicidal communication and announcements, or suicide attempts, without a sense of coping or hope.
   b. Coping: Personal stories about an individuals experience with suicidal thoughts or a suicide attempt, with a sense of hope, recovery, coping, or mentioning an alternative to suicide. The sentiment does not have to be positive. A neutral tone, or talking about difficult experiences with a sense of coping or mentioning recovery, is sufficient. Previous research suggests such messages may have a Papageno effect.
2. News reports about suicidal experiences and behavior except cases, often about celebrities:
   a. News suicidal ideation & attempts: about suicidal experiences without any mention of coping, including reports on suicidal ideation, suicide attempts, announcements of suicide, someone being put on "suicide watch", etc.
   b. News coping: about attempted or successful coping with or recovering from a suicidal crises
3. Tweets describing the experience of a person who lost someone to suicide in first or third person perspective:
   a. Bereaved negative: Describes the suffering or purely negative experience of a person who lost someone to suicide, including depression, grief, loss etc. These tweets necessarily refer to a suicide case, but are labeled as bereaved as long as they focus on the experience of bereaved individuals.
   b. Bereaved coping: Describes the experience of a bereaved person with a sense of hope, recovery or coping. The sentiment does not have to be positive. A neutral tone, or talking about difficult experiences with a sense of coping or mentioning recovery, is sufficient.
4. Reports of a particular completed or prevented suicide cases, often news reports:
   a. Suicide cases: About an individual suicide, or a timely or geographical suicide cluster. Suicide cases have priority over definition criteria of other categories (except tweets focusing on bereaved individuals which are always related to a suicide case). Previous research suggests such messages on individual suicide deaths (especially about celebrities) may have a Werther effect.
   b. Lives saved: News report or personal message about someone saving a life. In contrast to prevention tweets, these lives are often being saved coincidentally.
5. Calls for action: These are general statements calling for actions addressing the problem of suicide, and intending to spread problem awareness or prevention-related information.
   a. Awareness: Tweets intending to spread awareness for the problem of suicide, often focusing on high suicide rates or associations with bullying, racism, depression, veterans etc. without hinting at any solution. These are often reports of research findings or suicide statistics.

b. Prevention: Tweets spreading information about a solution or an attempt to solve the problem of suicide, including prevention at an individual (e.g., do not leave people alone in crisis situations) or public health level (e.g., safety nets on bridges). Hinting at a solution or a way of dealing with the problem is enough. No specific action needs to be described. These tweets often include a helpline number. Announcements of prevention events and broad recommendations for actions also count: donations, prayers with a focus on a solution for suicide, being there for someone, telling people that they matter, taking a course about suicide prevention, warning signs to watch out for.

6. Irrelevant: Messages that do not fit into any of the above categories:
   a. Suicide other: Anything about actual suicide, but not clearly related to any other above category, including murder-suicides, confident statements that something was not a suicide, convincing statements of not being suicidal, historical tweets about suicides that are a minimum of 40 ago (e.g., about Hitler's suicide), movies, books, novels, fiction about suicide, etc.
   b. Off-topic: messages that use the term suicide in a context other than suicide. This includes messages on euthanasia, suicide bombing, and suicide attacks, messages that are (suspected) jokes, irony, sarcasm, flippant remarks, or really unclear in terms of authenticity, messages that use suicidal/suicide to exaggerate an emotional experience (unclear if serious), or as a metaphor (e.g., political/financial/career suicide, suicide workout, suicidal immigration policies...), and messages about "suicidal animals" (e.g., killed by car).

## Analysis

Data analysis was done with R for inter-coder reliability, descriptive statistics and figures using R version 3.6.3. Main libraries used in R were the tidyverse, caret, and DescTools [33–35]. For training deep learning models, we used Python 3.6. Main packages were the ktrain wrapper [36] for the deep learning library TensorFlow Keras [37] and the scikit-learn library [38], for TF-IDF and SVM. For links to code and data, see the data and code availability statement.

### Text Preprocessing

We apply standard preprocessing strategies [see e.g., 39]: We replaced all URLs with a general marker token "*http*", all mentions (tags of Twitter users) with "*@user*", and lower-cased all words. The latter allowed using the smaller, more resource efficient BERT-lowercase model (see below). We kept Emoji, stopwords and punctuation, separated into single tokens, given that they can indicate the emotional connotation of a message (e.g., expressing excitement or surprise [40], or frequent singular pronouns indicating suicidal ideation [18]. We report the effects of additional standard different preprocessing steps, namely removing digits, punctuation, stopwords and lemmatization, in the SI. The basic preprocessing strategy yielded the most consistently high performance scores on the validation set, and is therefore used for all analyses. After preprocessing, the mean length of tweets in our labeled data set was 25 tokens, the 95th and 99th percentile 57 and 67 tokens (see Supplementary Figure S1). Based on this, we used 80 token's as the maximum sequence length for model input.

### Classification Tasks

***Task 1: Six main categories***. We trained our models to classify between categories with at least 200 tweets to allow for sufficient training data. From the categories of interest, these were messages about *(1) personal experiences of coping, (2) personal experiences of suicidal ideation and attempts, (3) suicide cases, (4) awareness,* and *(5) prevention.* We assigned all tweets from smaller categories (suicidal & coping news, negative & coping experiences of bereaved, lives saved) to the category *suicide other*, which belongs to the larger category of *irrelevant* tweets. In this task, we did not differentiate between irrelevant tweets that were about suicide (suicide other) and off-topic tweets, that use the word suicide in some other way. Instead, we subsumed suicide other and off-topic tweets in the category *(6) irrelevant.*

**Task 2: Detecting content about actual suicide.** This binary classification distinguishes between tweets that are *(1) about actual suicide* in the meaning of someone taking their own life, from tweets are *(2) off-topic*, i.e., use the word suicide in some other context. In our annotation schema, this task therefore separates the off-topic category from all other categories. The resulting label predictions allow to estimate the total volume of tweets about actual suicide, thereby improving on total volume estimates based on keyword searches.

**Machine Learning Models and Model Training**

Before training models, we divided the data set of 3200 tweets into a training (64%), validation (16%), and test set (20%), stratifying per tweet category to have a similar distribution in all sets. The training set was used for fitting the parameters of the classifier using 5-fold cross-validation. The validation set was used to tune the hyper-parameters (e.g., the learning rate) and to evaluate the model developed on the training data. After model training, we use the test set only once per model to estimate its ability to generalize to novel texts.

**Majority classifier.** We use a naive classifier that always predicts the majority class as a baseline to compare other models to.

**TF-IDF & SVM.** Term-frequency-inverse document frequency represents the text of tweets using weighted word *frequencies (f)*, which reflect how important a *term (t)* is to a *document (d, here a tweet)* in a corpus (all tweets). We slightly adjust the original formula for TF-IDF by adding 1 in the numerator and denominator, to ensure each word occurs at least once and prevent zero division[41]:

$$tf\text{-}idf(t, d)=tf(t, d) * \log((N+1)/(df + 1))$$

The resulting value increases proportionally to the number of times a word appears in the document and is offset by the number of documents in the corpus that contain the word. This helps adjust the weight of uncommon words which are more important for distinguishing different documents from each other than words which occur in every single document. After building the TF-IDF representation, we trained a Support Vector Machine classifier using all term values as features.

To identify the best TF-IDF representation and SVM classifier, we ran a grid search across the following dimensions: For TF-IDF, we (1) included only unigrams, or unigrams + bigrams, and (2) reduced the text to its n top features ordered by term frequency, where $n \in$ {10.000, 25.000, 50.000, N one}. For the SVM, we tested different hyper-parameters, namely, the (1) regularization parameter $C \in \{0, 1\}$, which determines the strength of the regularization, and (2) the class weight $cw \in$ {balanced, none}, which determines if the weights of the classes are automatically adjusted inversely proportional to class frequencies. We further tested (3) a linear and a radial basis function kernel, (4) decision function shapes one-vs-one and one-vs-rest. Optimal results were achieved including both unigrams and bigrams as text representation, 10.000 top features, and an SVM with C=0.82 in Task 1 and C=0.46 in Task 2, cw=balanced, a linear kernel, L2 penalty, and a one-vs-one decision function shape.

**BERT-base.** We used a transfer learning approach based on a pre-trained BERT (Bidirectional Encoder Representations from Transformers) base uncased model [42]. BERT is an auto-encoding, deep contextual language representation model developed by Google AI, which has 12 transformer layers, 12 self-attention heads and a hidden size of 768. It is designed to pre-train bidirectional representations of word sequences, that is, it learns from both the left-side and right-side context of a word in all of its layers. BERT was pre-trained with masked language modeling: a percentage (around 15%) of words in the sentence is randomly masked, and the model tries to predict the masked words from the sequence of other words. BERT was further trained to predict the next sentence from the previous sentence in the data.

**XLNet-base.** A known limitation of BERT is that it neglects dependence between the different masked words in a sentence. When predicting a word from a sequence that does not include the other masked words, BERT lacks information about the dependence between the masked words and the predicted word. Unlike autoregressive models, BERT further predicts all masked words simultaneously, and thus lacks some information about the order of words. XLNet [43] has a similar architecture as BERT, but addresses these shortcomings through permutation language modeling, predicting each word from all possible permutations of other words in the sentence. It thereby improves both on previous autoregressive models by using all words in the sentence, and on BERT by considering the order dependence of words. In addition, it incorporates some techniques from Transformers-XL [44] that also allows it to learn from the longer context before each word (relative positional encoding and segment recurrence mechanism).

**Fine-tuning of BERT and XLNet.** We fine-tuned the pre-trained BERT-base-uncased model and the XLNet-base model to our training data set as in Liu et al. (2019): we added one dense output layer to reduce the dimensions of the model's last layer to the number of labels in the classification task, and train all the parameters simultaneously. We ran a hyper-parameter search to determine good learning rate (LR) candidates and subsequently tested each LR by running three experiments with different seeds ∈ {1,2,3}. We aimed at finding the maximal LR associated with a still-falling loss (prior to the loss diverging), training for 5 epochs with learning rates ∈ {2e-5, 3e-5, 5e-5}. Reported results for BERT in Task 1 (6 classes) were the result of fine-tuning with a LR=2e-5 for 7 epochs and seed=1. Results for Task 2 (about actual suicide) are based on a BERT model trained with a LR=1e-5, 10 epochs and seed=1. Reported results for XLNet are based on model training with LR=2e-5, 8 epochs and seed=1 in both tasks.

**Metrics for Comparing Machine Learning Models**

We used various evaluation metrics to compare the different machine learning models. *Accuracy* indicates how many of all predictions (true positive and true negative) are correct. It is a global metric calculated over all classes in a data set. In data sets with large class imbalances, it can be high even if it always only predicts the majority class (e.g., the irrelevant category in Task 1). In this case, the model might not have learned anything despite high accuracy. *Precision* indicates the proportion of correct "positive" predictions out of all predictions, for example, how many of all predicted coping tweets were actually labeled as coping tweets by human raters. *Recall* indicates the proportion of all "true" cases (e.g., all actual coping tweets) that the model detects. The F1-score is the harmonic mean between Precision and Recall (F1=2*(precision*recall)/(precision+recall). Precision, recall and F1 score are calculated for each category and can be averaged across classes to produce a macro-average. For category-specific precision and recall, we provide 95% binomial confidence intervals calculated with the Clopper-Pearson method.

To compare models, we report macro-averages of model performance scores for both the validation and the test set, i.e., we calculate the mean of the performance measures of each class to have an aggregate measurement robust to class imbalance. While good scores on the training set indicate that the model has learned patterns existing in the training set, good scores on the test set additionally indicate an ability to generalize to novel data.

For determining the model to make predictions for a follow-up study [31], we decided a priori that we would prioritize precision over recall for task 1, which aims at identifying specific categories of tweets. The rationale behind this is that our follow-up studies focus on identifying specific Twitter signals (i.e., the percentage of coping tweets) that are associated with suicide cases and helpline calls. In such a situation, a false negative is less costly than a false positive, i.e., missing a tweet is less costly than falsely including a tweet into a certain category. Prioritizing precision ensures that we only count a tweet when it belongs to a category with a high probability. Furthermore, because of the large number of tweets, the proportion assigned to each category should accurately reflect the true proportion even if not all tweets are recognized. In contrast, Task 2 makes predictions that aim to capture the entire discussion about actual suicide on Twitter. When choosing the best model for Task 2, we focused on the F1-score. Here, we aim at capturing the total volume of tweets about suicide as fully as possible, as well as at accurate predictions at the tweet level.. False positives are less critical as a problem in Task 2 than in Task 1, because we look at total tweet volume and do not try to distinguish between the specific effect of a certain tweet category. This is best captured with the F1 score, which balances recall and precision. In any case, none of these a priori decisions at consequences for our results, given that BERT and XLNet performed very similarly, and much better than the other models.

## Evaluating Reliability and Face Validity of Model Predictions for BERT

### Comparing Model and Human Inter-Rater Reliability

To compare the models' reliability to human inter-rater reliability on novel data, we made predictions with one of the best models (BERT) for tweets from the full data set that were not part of the labeled data set. We selected 150 tweets per predicted label for each of the five relevant main categories. Two independent human coders manually labeled these tweets until we reached at least 80 per main category. The final set of 750 labeled tweets make up the reliability data set.

### Face Validity Check With The Predictions Data Set

For a face validity check, and a follow-up study [30], we estimated the daily volume of tweets per category that Twitter users may have been exposed to in between 1 January 2016 and 31 December 2018. For this, we created a data set with the same keywords and exclusion terms as the machine learning data set, but including retweets (to account for the full volume) and for a shorter time period of three years (determined by the follow-up study, see below). This resulted in English 7.150.610 tweets from users in the US. We used the machine learning model BERT to predict category labels for these tweets, and calculated the daily percent of tweets per category. We refer to this data set with model predictions for around 7 million tweets as the predictions data set. As a face validity check, we plotted the time-series of tweet volumes per category, and identified the events associated with the largest frequency peaks. For this, we investigated word frequencies on these days, read the tweets containing the most frequent terms, and googled these terms plus the date, or the tweet in quotes, to find (news) reports about the event. The follow-up study [30] investigates associations of these daily tweet volumes with suicide cases and helpline calls in the US. It has access to suicide case data from the Center for Disease Control and call data from the US suicide prevention Lifeline for the years 2016-2018, which was the reason to estimate tweet volumes for this time period.

Properties of all data sets used in the current study, including the labeled machine learning data set, and those for comparing model and human performance, and the face validity check, are depicted in Figure 2.

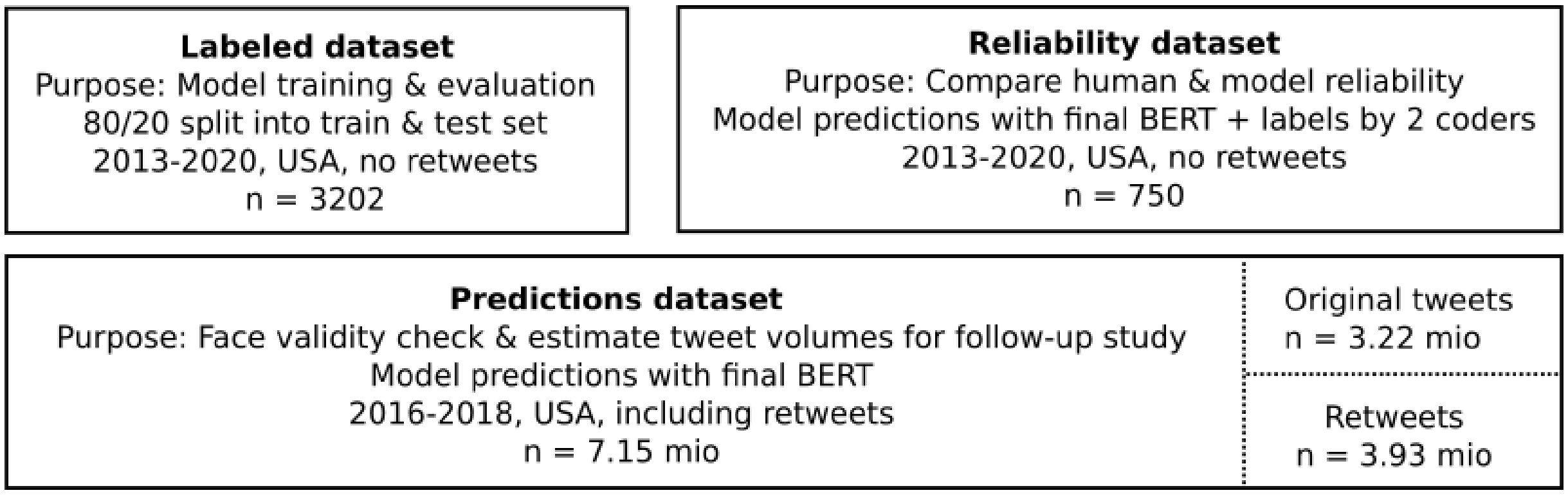

**Figure 2.** Overview of characteristics of data sets. Each box describes the purpose of the data set, further details on how it was used or created, and the sample size. Only the predictions data set includes retweets, as it aims to capture the full volume of tweets posted on a given day.

# Results

## Frequency of Tweets per Category

Table 2 displays the proportion of tweets per main category in our labeled data set, and in two different samples used to estimate the natural frequency of categories on Twitter. First, we used a sub-sample of the labeled data set of 1000 randomly selected tweets (i.e., selected without key words or model predictions, 996 after 4 labeling, see section *Creating the Annotation Scheme and Labeled Data Set*) to estimate the frequency of original tweets, without counting retweets. For the second estimate, we used predictions by the best model (BERT) to label tweets in the predictions data set, which includes retweets. Here, we adjusted the frequency per category by dividing by the model's recall (i.e., the proportion of true cases the model detects). The percentage for the irrelevant category was calculated by subtracting the percentages of relevant categories from 100. The two estimates were similar for suicidal ideation & attempts and suicide cases.

The percentages per category in Table 2 demonstrate that we managed to include proportionally more of the rare tweet categories like coping and suicidal ideation stories in our training set. Nevertheless, irrelevant tweets, in particular off-topic tweets, still make up a majority of tweets in our data set.

**Table 2. Distribution of tweets across categories for manual labels and model predictions.**

| Category label | | Total labeled sample n=3202 % (n) | Subset of labeled tweets, randomly selected n=1000 | Estimated frequency in predictions data set (including retweets) n=7.15 million | |
|---|---|---|---|---|---|
| | | | | ***Task 1*** | ***Task 2*** |
| Suicidal | | 8.87% (284) | 6.33% | 5.13% | |
| Coping | | 6.5% (205) | 2.71% | 1.26% | |
| Awareness | | 9.81% (314) | 12.54% | 22.06% | |
| Prevention | | 14.27% (457) | 7.13% | 15.51% | |
| Suicide cases | | 16.05% (514) | 12.95% | 16.16% | |
| Irrelevant | | 44.5% (1428) | 58.33% | 39.88% | 76.52% |
| **Sub-categories of irrelevant** | News suicidal | 2.12% (68) | 2.01% | | |
| | News coping | 0.84% (27) | 0.5% | | |
| | Bereaved negative | 1.06% (34) | 0.7% | | |
| | Bereaved coping | 1.06% (34) | 0.5% | | |
| | Live saved | 0.41% (13) | 0.2% | | |
| | Suicide other | 13.74% (440) | 20.68% | | |
| | Off-topic | 25.36% (812) | 33.73% | | 23.48% |

## Model Performance

Given that the performance of both deep learning models with fixed seeds and parameters varies a bit from run to run due to internal segmentation, we ran these models five times. We report the average of all five runs in this section and include the metrics for each individual run in the SI (see Table S2 to S5).

### Task 1: Six Main Categories

Performance scores averaged across all six tweet categories (Table 3) show that all deep learning models performed very similarly, and substantially better than the TF-IDF & SVM approach. Yet, TF-IDF & SVM was clearly better than a naive majority classifier. It reached scores from 0.61 to 0.66, which were nearly identical on the validation and test set. For BERT and XLNet, all scores were at or above 0.70, and only 0.1-0.3 lower on the test- than the validation set, indicating good ability to generalize to new tweets. Macro-average performance scores in all five runs for BERT and XLNet were about 10% higher than TF-IDF & SVM macro-averages (Table S2).

**Table 3. Macro-averaged performance metrics and accuracy cross all 6 categories on the validation and test set.**

| | Validation set n=513 | | | | Test set n=641 | | | |
|---|---|---|---|---|---|---|---|---|
| **Model** | **Pr**[a] | **Re** | **F1** | **Acc** | **Pr** | **Re** | **F1** | **Acc** |
| Majority classifier | 0.07 | 0.17 | 0.10 | 0.45 | 0.07 | 0.17 | 0.10 | 0.44 |
| TF-IDF & SVM | 0.61 | 0.63 | 0.62 | 0.66 | 0.61 | 0.65 | 0.62 | 0.66 |
| BERT | 0.73 | 0.71 | 0.71 | 0.76 | 0.72 | 0.69 | 0.70 | 0.73 |
| XLNet | 0.74 | 0.73 | 0.73 | 0.77 | 0.71 | 0.71 | 0.71 | 0.74 |

[a] Pr=Precision, Re=Recall, Acc=Accuracy.

[b] For BERT and XLNet, metrics are averages of 5 runs (separate runs in SI).

Given that the macro-average performances were substantially lower for the majority classifier, we focus on the three other models for intra-class scores (Table 4 and Figure 3a). To choose a model for making predictions, we focused on F1 scores and precision (see section on evaluation metrics). F1 scores were higher for BERT and XLNet than TF-IDF & SVM for all relevant categories, with clear differences for some categories (suicidal, coping, awareness), and very small differences for others (suicide cases, prevention). For BERT and XLNet, F1-scores were almost identical for all categories. BERT yielded higher precision for coping and and prevention tweets, two crucial categories for a follow up publication [31]. We therefore chose BERT as the model to make predictions for further analyses. It should be noted here that confidence intervals are quite large due to the limited size of the test set per class, and entirely overlap for BERT and XLNet, and somewhat overlap for most categories with TF-IDF & SVM. Nonetheless, performance scores in the five runs of BERT and XLNet were higher than TF-IDF & SVM in almost all cases for all relevant categories (Table S3). Only in the case of precision for prevention tweets, TF-IDF & SVM performed similarly well in three out of five runs.

**Table 4. Intra-class performance metrics for deep learning models on the test set.**

| Test set | Suicidal ideation & attempts n=57 | | | Coping n=42 | | |
|---|---|---|---|---|---|---|
| **Model** | **Precision** | **Recall** | **F1** | **Precision** | **Recall** | **F1** |
| TF-IDF&SVM | 0.32 [21.93-43.58] | 0.44 [30.74-57.64] | 0.37 | 0.44 [31.55-57.55] | 0.64 [48.03-78.45] | 0.52 |
| BERT[a] | 0.58 [43.25-73.66] | 0.45 [32.36-59.34] | 0.51 | 0.76 [59.76-88.56] | 0.69 [52.91-82.38] | 0.72 |
| XLNet[a] | 0.60 [46.11-74.16] | 0.54 [40.66-67.64] | 0.55 | 0.71 [54.80-83.24] | 0.74 [57.96-86.14] | 0.73 |
| | **Awareness n=63** | | | **Prevention n=91** | | |
| **Model** | **Precision** | **Recall** | **F1** | **Precision** | **Recall** | **F1** |
| TF-IDF&SVM | 0.65 [51.60-76.87] | 0.62 [48.80-73.85] | 0.63 | 0.83 [74.00-90.36] | 0.82 [73.02-89.60] | 0.83 |
| BERT | 0.71 [58.05-81.80] | 0.70 [56.98-80.77] | 0.70 | 0.81 [71.93-88.16] | 0.89 [80.72-94.60] | 0.85 |
| XLNet | 0.69 [56.74-79.76] | 0.74 [62.06-84.73] | 0.72 | 0.82 [72.27-88.62] | 0.87 [78.10-93.00] | 0.84 |
| | **Suicide cases n=103** | | | **Irrelevant n=285** | | |
| **Model** | **Precision** | **Recall** | **F1** | **Precision** | **Recall** | **F1** |
| TF-IDF&SVM | 0.70 [60.82-78.77] | 0.74 [64.20-81.96] | 0.72 | 0.74 [67.78-79.18] | 0.63 [57.27-68.77] | 0.68 |
| BERT | 0.75 [65.14-82.49] | 0.77 [67.34-84.46] | 0.76 | 0.64 [57.76-69.11] | 0.65 [59.06-70.45] | 0.64 |
| XLNet | 0.78 [68.31-85.52] | 0.75 [65.24-82.80] | 0.76 | 0.68 [61.96-73.46] | 0.64 [57.99-69.44] | 0.66 |

[a] Scores are averages across 5 model runs for BERT and XLNet, see Appendix 1, Table S3 for separate runs.

Overall, BERT correctly classified 73% of tweets in the test set. F1-scores lay between 0.70 to 0.85 for the different categories of interest (see Table 4 and Figure 3a), with the exception of the suicidal ideation & attempt category with an F1-score of 0.51. More specifically, recall for suicidal ideation & attempt was relatively low (0.45), indicating difficulties in detecting all such tweets, whereas precision was higher with 0.58. All performance scores were particularly good (>0.81) for prevention tweets, and also quite high for tweets about suicide cases (>0.75). For coping tweets, BERT achieved very high precision (0.76), but lower recall (069), which resembles the pattern observed for suicidal tweets. Performance scores for awareness tweets were around 70%.

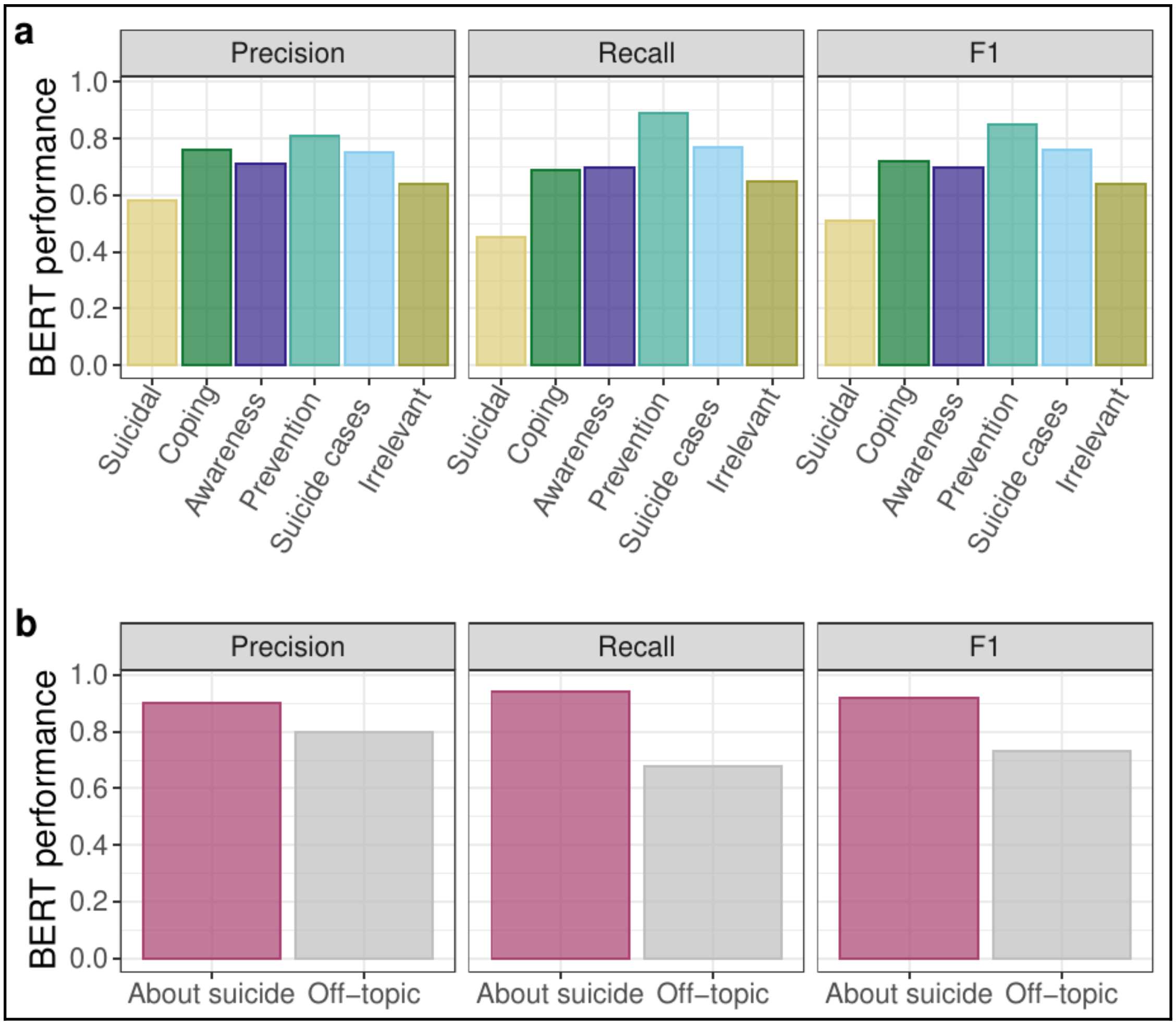

**Figure 3. Performance scores per category for BERT.** (a) For the 6 main categories and (b) for tweets about actual suicide vs. off-topic tweets.

**Task 2: About Actual Suicide**

Best performances for separating tweets about actual suicide from off-topic tweets (Table 5) were observed with BERT. However, XLNet performances were very similar with largely overlapping confidence intervals. With TF-IDF & SVM, recall for about suicide tweets and precision for off-topic tweets were significantly lower than the deep learning scores, whereas precision for about suicide and recall for off-topic were not significantly different. The model with overall highest scores, BERT, correctly labeled 88.5% of tweets as about suicide vs. off-topic with very similar scores on the validation and test set. F1-scores for about suicide vs. off-topic tweets in the test set were 0.92 and 0.73, respectively (Table 6). All metrics were at least 10% higher for tweets about suicide than off-topic tweets. In particular, recall was very high for tweets about suicide (94%), which indicates that volume estimates for tweets relating to suicide will be quite complete. Precision for tweets about suicide was 90%, underlining that positive predictions of the model were very reliable.

**Table 5. Macro-averaged performance metrics and accuracy for Task 2 (about suicide vs. off-topic) on the validation and test set.**

| | Validation set n=513 | | | | Test set n=641 | | | |
|---|---|---|---|---|---|---|---|---|
| **Model** | **Pr[a]** | **Re** | **F1** | **Acc** | **Pr** | **Re** | **F1** | **Acc** |
| Majority classifier | 0.37 | 0.50 | 0.43 | 0.75 | 0.37 | 0.50 | 0.43 | 0.75 |
| TF-IDF & SVM | 0.74 | 0.77 | 0.75 | 0.80 | 0.75 | 0.77 | 0.76 | 0.81 |
| BERT | 0.85 | 0.81 | 0.83 | 0.88 | 0.85 | 0.81 | 0.83 | 0.88 |
| XLNet | 0.84 | 0.78 | 0.81 | 0.87 | 0.83 | 0.80 | 0.81 | 0.87 |

[a] Pr=Precision, Re=Recall, Acc=Accuracy.

**Table 6. Intra-class performance metrics for deep learning models in Task 2 (about suicide vs. off-topic) on the test set.**

| Test set | About suicide n=478 | | | Off-topic n=163 | | |
|---|---|---|---|---|---|---|
| **Model** | **Precision** | **Recall** | **F1** | **Precision** | **Recall** | **F1** |
| TF-IDF & SVM | 0.89 [85.74-91.71] | 0.85 [80.96-87.64] | 0.87 | 0.60 [53.03-67.49] | 0.69 [61.63-76.30] | 0.65 |
| BERT[a] | 0.90 [87.42-92.81] | 0.94 [91.64-96.07] | 0.92 | 0.80 [71.62-85.67] | 0.68 [60.35-75.17] | 0.73 |
| XLNet[a] | 0.90 [87.12-92.59] | 0.93 [90.68-95.38] | 0.92 | 0.76 [68.60-83.06] | 0.67 [59.72-74.60] | 0.71 |

[a] Scores are averages across 5 model runs for BERT and XLNet, see Appendix 1, Table S5 for separate runs.

## Comparing Model and Human Inter-Rater Reliability

### Task 1: Six Main Categories

The inter-rater reliability (Cohen Kappa) for the six main categories was κ 0.70 (95% CI 0.67-0.74). between two human coders, and κ 0.60 (95% CI 0.56-0.64) and κ 0.63 (95% CI 0.59-0.67) between each human coder and the BERT model, respectively. Lower agreement with BERT compared to between humans was mainly driven by the irrelevant class. Excluding it from analysis yielded κ 0.85 (95% CI 0.82-0.89) between human raters, and κ 0.81 (95% CI 0.77-0.85) and 0.80 (95% CI 0.76-0.84) between BERT and each human rater. These overlapping confidence intervals indicate a non-significant difference, and show that BERT achieved near human-level accuracy for the relevant categories.

Precision and recall comparisons between model and human performance per tweet category are shown in Table S6, and the confusion matrix for coder 1 and BERT in Figure 4. We first report metrics for the model versus each coder, with the coder as the ground truth. Second, we report the same metrics for coder 2 compared to coder 1 as the ground truth. Model precision is clearly lower than between-human precision for *suicidal ideation & attempts* and *awareness* messages, more comparable for *coping stories*, and very similar for *prevention* and *suicide case tweets.* Recall is clearly higher between human raters for *suicidal and coping* stories, and similar for *suicide cases*. For awareness and prevention tweets, the model actually achieves better recall than human coders. The model seems thus quite good at detecting awareness tweets, but is not very precise in return.

### Task 2: About Actual Suicide

When categorizing tweets as being about actual suicide vs. off-topic, human inter-rater reliability was κ 0.44 (95% CI 0.29-0.58), compared to κ 0.15 (95% CI -0.07-0.37) and κ 0.21 (95% CI -0.01-0.44) between each coder and BERT. These low Kappa coefficients were mainly driven by low performances for the irrelevant off-topic category between both human coders (coder 1 – coder 2: precision=0.52, recall=0.44, F1: 0.48), which were even lower when comparing human to model labels (coder 1 - BERT: precision=0.26, recall=0.13, F1=0.17; coder 2 - BERT: precision=0.39, recall=0.16, F1=0.23). In contrast, performance for the about suicide category was very high when comparing human labels (precision=0.96, and recall=0.97, F1=0.96), as well as when comparing human and model labels (coder 1 - BERT: precision=0.94, recall=0.98, F1=0.96, coder 2 - BERT: precision=0.94, recall=0.98, F1=0.96). This shows that the two coders and the model agreed which tweets were about actual suicide, and detected most tweets the other coder had labeled as about suicide. Yet, they agreed less when judging whether a tweet was not about actually suicide, hinting at the inherent difficulty of judging whether something is serious, sarcastic, a metaphor or not. In any case, for future studies correlating tweets about suicide with behavior in the population, only the about suicide category, which can be reliably detected by humans and the machine learning model, is relevant.

## Error Analysis

Figure 4 shows the confusion matrix of true and predicted labels for BERT for the six main categories in the reliability data set. Most misclassifications were predictions of the label irrelevant. Such false negatives are less problematic than misclassifications between relevant categories, as we prioritized precision over recall. Between relevant categories, there were five cases in which coder 1 and the model labeled more than 9 but max. 15% of tweets differently: (1 & 2) confusions between coping and suicidal tweets in both directions, (3 & 4) confusions between awareness and prevention tweets in both directions, and (5) tweets about suicide cases misclassified as awareness tweets.

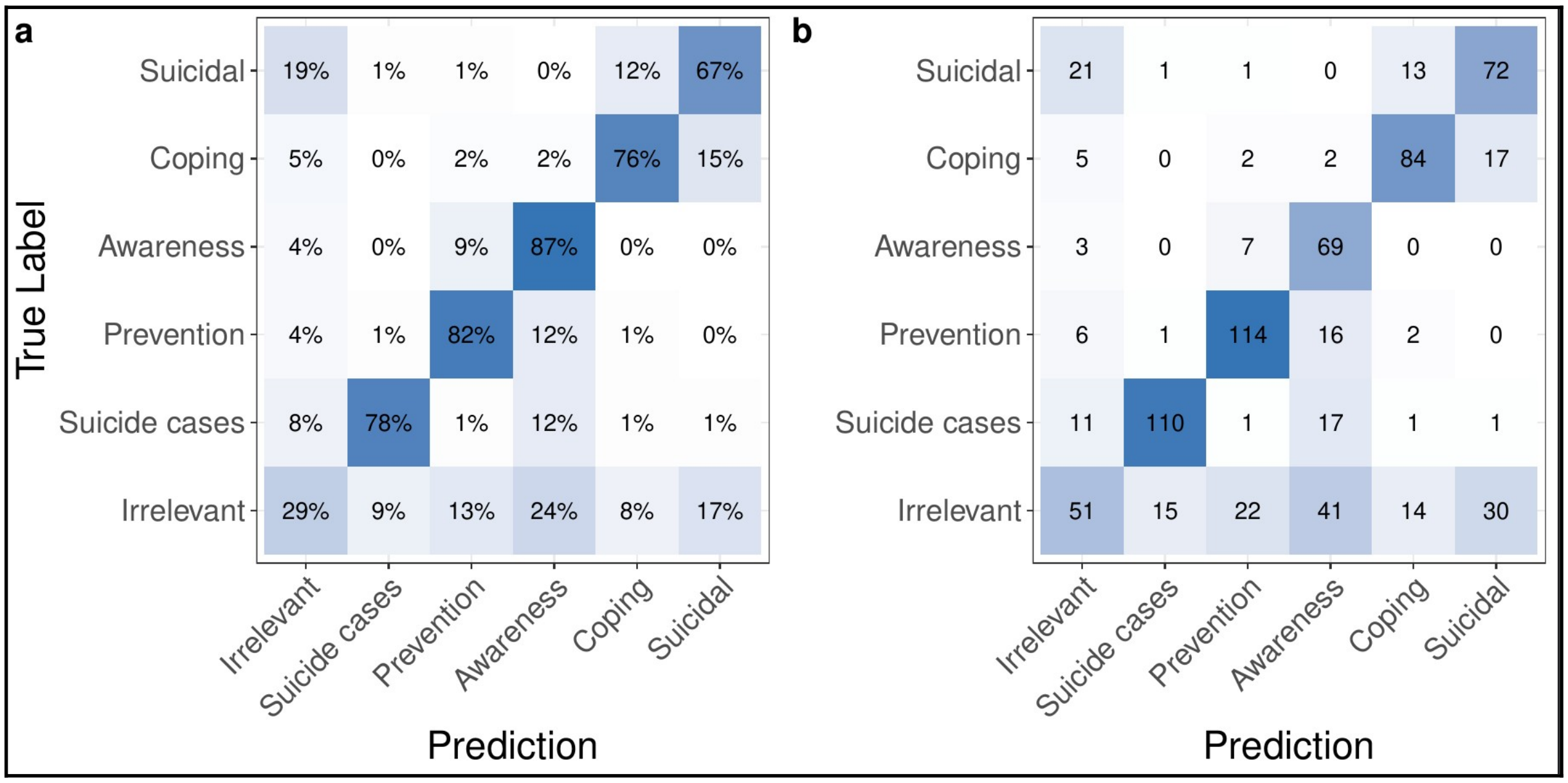

**Figure 4. Confusion Matrix of true and predicted labels in the reliability data set.** (a) percentages and (b) count of tweets per true and predicted category. The diagonal from bottom left to top right represents correct predictions. True labels are labels by coder 1, and predicted labels by BERT.

In case 1 and 2, coder 2 classified about a third of tweets in the same way as the model, indicating the difficulty of clearly separating personal stories about suicidal ideation and coping. In particular, 4 out of 13 tweets labeled as suicidal by coder 1 (12% of all suicidal tweets in the test set), were labeled as coping by both the model and coder 2. The model's label more closely matched the category definition in at least 5 of the 13 cases. Only 2 of the model predictions were clear errors, all other tweets were ambiguous. Many of these ambiguous tweets described suicidal ideation in the past, implicitly hinting that the suicidal phase was over at the moment the tweet was written. For misclassifications of coping tweets as suicidal (15% of coping tweets), coder 1 and 2 also disagreed in 6 out of 17 cases, suggesting that many of these misclassified tweets were ambiguous. Although 12 of these 17 misclassifications were actual model errors, most of them were understandable, given that coping was described implicitly by means of suicidal ideation in the past, or that strong suicidal ideation was expressed along with a way in which the person deals with it.

Misclassifications of awareness as prevention tweets (12%) were errors by coder 1 rather than the model in 4 of 7 cases, indicating that model performances are higher for awareness tweets as scores in Table 3 suggest. In contrast, when the model labeled prevention tweets as awareness, these were mostly clear mistakes (only 3 out of 16 were ambiguous errors by the coder). Finally, suicide cases mislabeled as awareness were mostly actual errors by the model, but ambiguous in 4 and actually correct in 2 out of 17 cases.

There was further a strong confusion between stories of suicidal ideation and one particular type of irrelevant tweet: tweets manually labeled as not-serious or unclear if serious (dimension 3). Of all tweets predicted to be suicidal stories, 11% were not serious/unclear, compared to only 1-3% in the other predicted categories. Of all non-serious/unclear tweets, 45% were correctly classified as irrelevant, 32% were wrongly classified as suicidal compared to 2-10% wrongly assigned to the other categories. The 13 non-serious tweets misclassified as suicidal included 4 exaggerations, 2 sarcastic remarks, 2 tweets with song lyrics or band names with the terms suicidal ideation/thoughts, 1 metaphoric use, and 2 statements about not being suicidal.

### Face Validity Check with Daily Time Series Peaks per Category

Figure 5 illustrates the daily percent of tweets in each predicted category in the predictions data set. As a face validity check, we identified the events that were mainly associated with each of the largest peaks in the time series of tweets per predicted category. For coping, prevention and suicide case tweets all highly frequent tweets were correctly classified tweets. Both highly shared coping tweets were from individuals who had survived a suicide attempt. Prevention peaks were related to the yearly World Suicide Prevention Day, and to increased prevention efforts around Christmas and the New Year, and to increased lifeline calls after Trump's election [45]. All identified peaks of tweets about suicide cases were related to actual instances of someone taking their own life. For awareness tweets, all but one peak were driven by actual awareness tweets. This single tweet (labeled Same-sex marriage in Figure 3) was ambiguous, as it reported a research finding like a typical awareness tweet, but the finding was somewhat prevention related. Most awareness peaks were driven by tweets that cite a statistic about suicides. Of the tweets driving the three largest peaks in the suicidal category, only one was clearly suicidal ideation, another a somewhat cynical tweet about someone else wanting to commit suicide, and a third was a clear confusion with an actual coping tweet. This face validity check thus reflects the high precision of the model for prevention, awareness and coping tweets, as well as the lower performance for suicidal ideation & attempt tweets.

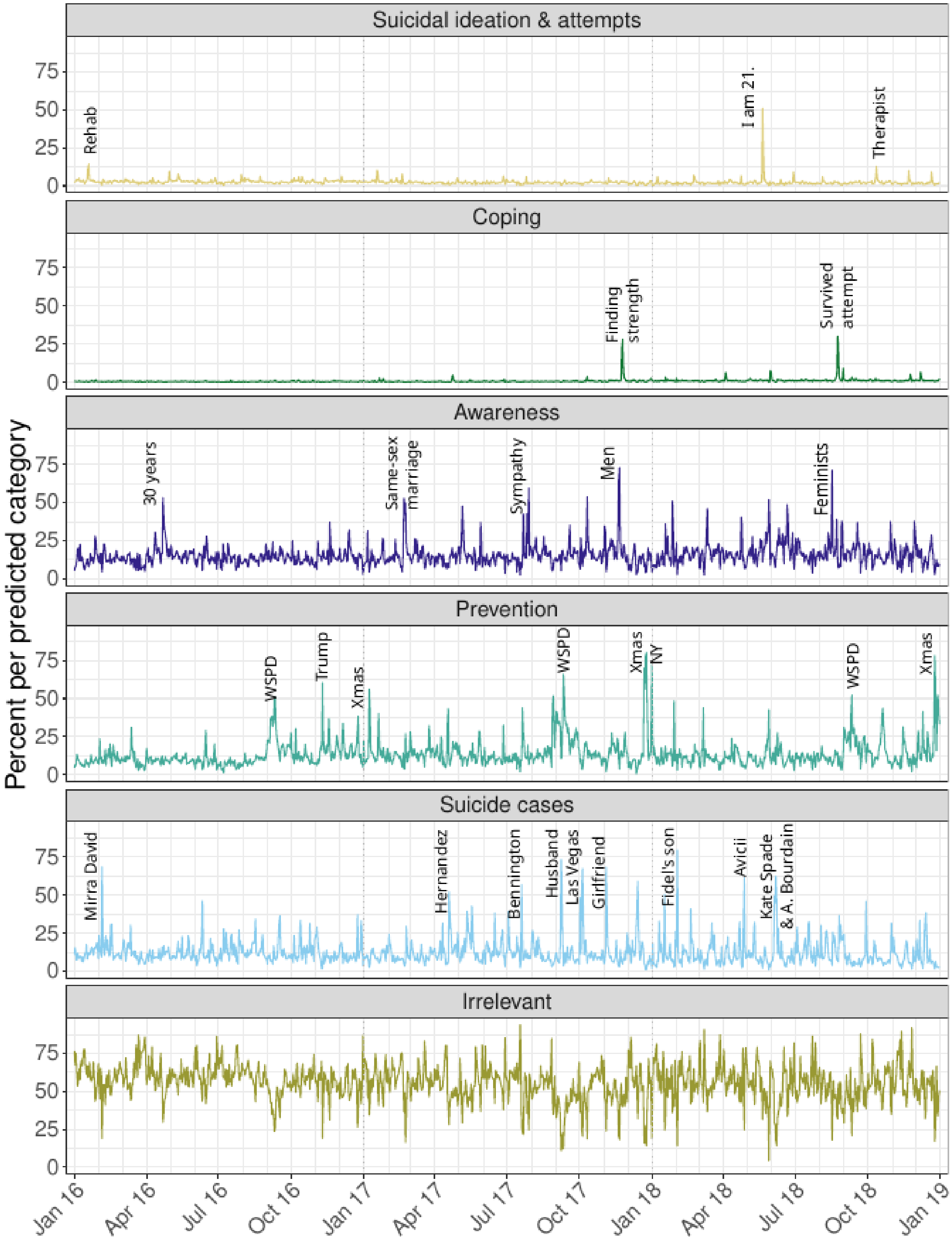

**Figure 5. Daily percent of tweets per predicted category in the predictions data set (n=7.15 million).** The daily value subsumes original and retweets per category. Keywords for event peaks: *Suicidal ideation & attempts:* Rehab: 4000 retweets of a rather sarcastic tweet from someone getting "punished" in rehabilitation because he said "fucking lit" that another patient was about to commit

suicide. I am 21: A personal story retweeted 2200 times, describing someone's successful journey from a difficult childhood, through a long suicidal crisis, to a University degree and a full time job, falsely labeled as suicidal, as it describes not just a suicidal crisis, but also coping. Therapist: 900 retweets of someone reporting no suicidal thoughts as their therapist asked, although they have them very frequently. *Coping*: Finding strength: Around 6000 retweets of a story of someone finding strength three weeks after a suicide attempt. Survived attempt: A marine corps veteran with PTSD tweets about his survived suicide attempt, around 5000 retweets. *Awareness*: Sympathy: A tweet saying people who died by suicide need care while still alive, rather than sympathy when they are dead, retweeted 3500 times. Men: 7000 retweets of a tweet mentioning that suicide is the largest cause of young men. Feminists: 6000 retweets of a tweet saying that feminists who want equality should also consider that boys are double as likely to die from suicide than girls. 30 years: retweets and discussion of a federal data analysis results that suicide in the United States had risen to the highest levels in nearly 30 years. Same-sex marriage: Many tweets on research finding that suicide rates drop after legalization of same-sex marriage. *Prevention*: Trump: Increased calls to suicide hotline after Trump's election. WSPD: World suicide prevention day. Xmas: Christmas. NY: New Year. *Suicide cases*[46]**:** Aaron Hernandez (American football player). Chester Bennington (singer of Linkin Park). Husband: retweets of tweet by a woman remembering her husband's suicide. Las Vegas: Many retweets of a reply correcting a tweet by Trump, by stating that the shooter killed himself. Girlfriend: Many retweets of a tweet about a girlfriend that killed herself. Fidel's son: Fidel Castro Diaz-Balart.

## Discussion

Due to the effort required for manual annotation of texts, previous research on media and suicide prevention was limited by small sample sizes, or by data sets put together using keyword search. Keywords either capture only a particular type of text (e.g., containing celebrity names) or lump together a variety of different texts that contain broad search terms (e.g. "suicide") [see e.g., 2,10–12] Additionally, research on the correlation of social media content with suicide cases in the population remains extremely scarce [9–13]. The current study extends media research on suicide prevention by focusing on a broad range of suicide-related content on social media, and by developing a reliable and efficient content labeling method based on machine learning, enabling fine-grained analysis of large data sets. We first developed a comprehensive annotation scheme for suicide-related content that includes new content types more typical on social than traditional media, such as personal stories of coping or suicidal ideation, or calls for action addressed at follower networks. Based on this systematic labeling scheme, we then tested the ability of different machine learning algorithms to distinguish five content types that seem particularly relevant based on previous research [e.g., 4,2,30,6]. We further applied these methods to separate tweets about actual suicide, i.e., in the meaning of someone taking their own life, from tweets that use the word suicide in some other way or context (binary classification). Our results for these two classification tasks show that machine learning methods, in particular deep learning models, achieve performances comparable to both human performance and to state-of-the-art methods in similar tasks [16,47].

Our study is one of the first to automatically classify social media data other than suicidal ideation into categories relevant for suicide prevention. Only three studies, two of which used the same data set, have previously applied machine learning to distinguish specific types of social media postings other than suicidal ideation [9,16,27]. We extend these studies in several ways: Rather than classifying emotions in tweets about specific celebrities [9] or using a relatively small set of 816 tweets put together with a focus on suicidal ideation and celebrity names [16,27], we trained models to categorize any type of tweets containing suicide-related terms in a much larger data set than in previous studies. Further, our larger data set enabled us to use deep learning models that can account for differences in the meaning of words across contexts, rather than only considering word frequencies. Finally, our annotation scheme introduces more fine-grained and particularly prevention-relevant categories.

Specifically, it includes personal coping stories, for which some research on traditional media suggests preventive effects [4–6], and we distinguish awareness from prevention-focused tweets [28].

## Principal findings

Regarding the machine learning results, pre-trained deep learning models fine-tuned to our data clearly outperformed a naive majority classifier and a linear Support-Vector-Machine classifier based on the word frequency representation TF-IDF. BERT and XLNet achieved F1-scores of 0.70 and 0.71 in the six-category classification and 0.83 and 0.81 in the binary about suicide vs. off-topic classification in the test set. These scores were only slightly lower or even identical to those in the validation set, indicating good generalization to novel data. The clear advantage of deep learning models over TF-IDF & SVM suggests that there is crucial information about meaning in the context of words beyond what mere word frequencies can capture in tweets about suicide. While deep learning models were better than the more traditional approaches, the performance of BERT and XLNet was very similar. Advantages of XLNet over BERT include its ability to learn from long contexts and to consider dependencies between all words in the sentence. It seems that these advantages cannot fully play out given the limited number of words in tweets.

The six investigated tweet categories separated five important categories, including personal stories about either (1) coping or (2) suicidal ideation and attempts, calls for action that spread (3) problem awareness or (4) prevention-related information and (5) tweets about suicide cases, from other tweets (6) irrelevant to this categorization. Performance scores per category were nearly indistinguishable for BERT and XLNet. Which model performed better depended on the metric and the category, and varied between model runs. BERT had slightly higher precision than XLNet for two important categories for a follow-up publication [31], and was therefore chosen as the model to make predictions and test reliability. Although our data set included a much broader set of tweets than previous studies focusing on similar prevention-related tweet categories [16,27], our machine learning performances were comparable or better than in these previous studies, with the exception of the suicidal ideation category.

In general, BERT and XLNet were better at classifying tweets that are also easier to distinguish for humans, including more homogeneous classes like prevention and suicide cases. These often included similar keywords, such as prevention, hotline, lifeline or the phrase “committed/commits suicide” (see word clouds in SI). For these categories, BERT performance was very similar to human inter-rater performances. BERT and human performance were also comparable for coping stories. Only for more subjective classes, like suicidal ideation & attempt stories, the model’s performance was lower. The error analysis suggests confusions with sarcastic, joking, exaggerated or metaphoric uses of the word suicide as one part of the explanation. Such non-serious messages are hard to distinguish from genuine suicidal ideation for both the model and humans. The gap between human and model performance is largest for suicidal tweets, hinting that this distinction is even harder for the model. Both humans and the model miss many ambiguous expressions of suicidal ideation (low recall). In contrast, between-human precision is much higher than the model’s precision. This, shows that there are many suicidal ideation tweets that humans can clearly identify as such, while model reliability can still be improved for these types of tweets.

The analysis of the most common model errors demonstrates that its mistakes were mostly not trivial. Most confusions of suicidal and coping tweets by the model were tweets in which the feelings of the tweet author were actually quite ambiguous. This suggests that much higher performance scores are hard to reach for these personal stories about suicide. Nonetheless, for tweets about suicidal ideation in the past, which implicitly express that coping occurred, there may be room for improvement through adding more training examples. Furthermore, the error analysis suggests possible improvements for tweets about prevention and suicide cases. In contrast, the model actually helped detect errors by the human coder for awareness tweets labeled as prevention.

When distinguishing tweets about actual suicide from off-topic tweets, the model achieved excellent performance scores in particular for tweets about actual suicide, with no difference between the two deep learning models. In other words, tweets labeled as “about suicide” are reliably actual tweets about suicide, and most such tweets are detected by the models. The use of any of these models for future research is thus very promising.

Using the final BERT models for both classification tasks, we estimated the percentage of tweets per category out of all suicide-related tweets in the US in the years 2016-2018. Overall, about 6% were personal stories of concerned individuals, with about 5% on suicidal ideation or attempts, and about 1% on coping stories. Estimates for awareness, prevention and suicide case tweets were around 22%, 16% and 16% of tweets, respectively. We plotted the daily volume per tweet category and investigated tweets during peaks in the time-series. Most of these tweets were correctly classified by our models, and peaks often coincided with events matching the particular category (e.g., the world suicide prevention day, or a celebrity suicide), which highlights the face validity of our model predictions. Finally, around three quarters of all suicide-related tweets actually referred to someone taking their own life, while the rest used the term in another meaning or context (e.g., euthanasia, suicide bombers, jokes, metaphors exaggerations).

## Limitations and Future Work

Despite our data set being more comprehensive than any existing data set on the topic, one of the limitations of our study is still the size of the training data set, which is crucial for training deep learning models. This concerns in particular the rarer categories which we have not yet used for machine learning in the current study (e.g., bereaved experiences, lives saved). The data set could further benefit from adding more examples of coping messages that describe suicidal ideation and behavior in the past, thereby implicitly indicating coping (see category definition in Appendix 2). Furthermore, some tweets in the categories suicide other and off-topic might warrant to be investigated separately, given recent findings of possible protective effects of flippant remarks and humor, or negative portrayals of suicide in the form of murder suicides [11]. Similarly, the suicide case category may warrant being separated into suicide news and condolence messages, which may have protective effects [9], and tweets about suicide cases may warrant filtering out those about celebrities [2]. Higher classification performance for the category suicidal ideation in Burnap et al. [16] shows that a focus on this category during data collection could improve our model.

Finally, a number of limitations apply to automated text analyses, like machine learning, in general. First, there are no traces of images, videos or the content of the URLs shared in the text of tweets, although this additional information can crucially affect the meaning of a tweet. Second, some things are only implicitly expressed or very subjective, and thus hard to capture with such methods, but also hard to reliably recognize for humans. For instance, it is hard to clearly differentiate coping from suffering even for humans, who have some knowledge about how such experiences look like in the real world. It is even more difficult to capture such subjective experiences using word frequencies. Deep learning models such as BERT and XLNet, having been trained on huge amounts of text produced by humans, may be able to capture some of these nuances, but need large amounts of training examples. Third, a machine learning model can only recognize example tweets that are similar enough to the examples in the training set, and only predict the predefined categories. In contrast, a human coder might recognize new ways of expressing the same meaning, or the need to introduce a new category. We partially addressed the latter limitation through our extensive labeling process, ensuring that we captured all typical message categories by including a random set of tweets.

Nonetheless, including more and different examples for suicidal ideation and coping stories in future studies would likely improve model performance.

### Conclusion and practical implications

The field of media and suicide research has only recently started to evolve to consider social media content as relevant in the assessment of media effects. The current work makes two major contributions to this field: First, it provides a systematic overview of different content types that are common on social media, which may be useful as a content labeling scheme for future research on the topic. Some of the categories identified have been found to be relevant to suicide prevention particularly in other media types. For social media content, these associations with indicators of behavior, particularly suicidal behaviors and help-seeking, remain to be tested accordingly. Second, the machine learning methods enables researchers to assess large amounts of social media data, and subsequently correlate it with available behavioral data of interest, e.g., suicides or help-seeking data. In this way, this work enables systematic large scale investigations of associations between these behaviors and fine-grained message characteristics of social media postings [see e.g., 31]. Such large scale investigations will contribute to accumulating robust evidence on which characteristics are actually harmful and protective. Furthermore, future applications of the developed models might include the screening of social media content to detect other types of content associated with suicide cases that have not been described in previous research. The classification performances of the developed models demonstrate the strong potential of machine learning, in particular deep learning, for media suicide effects research.

### Acknowledgments

This work was supported by the project V!brant Emotional Health grant “Suicide Prevention media campaign Oregon” to T. Niederkrotenthaler, and Grant No. VRG16-005 from the Vienna Science and Technology Fund to D. Garcia. We thank Konstantin Hebenstreit for help with data collection and cleaning.

### Data and code availability

The data sets generated during and analyzed during the current study are available on GitHub. The code and data for the statistical analyses and figures (inter-coder reliability, frequency of categories) as well as confidence intervals for machine learning is available at https://github.com/hannahmetzler/TwitterSuicideR. Code for training the machine learning models is available at https://github.com/HubertBaginski/TwitterSuicideML. Raw data only includes tweet IDs, not tweet text, to protect the identity of the tweet authors. Tweets can be re-downloaded via the Twitter API using these Ids.

### Conflicts of Interest

The authors report no conflict of interest.

### Multimedia Appendices

Appendix 1: Supplementary Information

Appendix 2: Annotation scheme

## Abbreviations

Acc: Accuracy

BERT: Bidirectional Encoder Representations from Transformers

Pr: Precision

Re: Recall

SVM: support vector machine

TF-IDF: Term frequency - inverse document frequency

# Appendix 1: Supplementary Information

# Detecting Potentially Harmful and Protective Suicide-related Content on Twitter: A Machine Learning Approach

**Authors**: Hannah Metzler[1,2,3,4,5], Hubert Baginski3[,6], Thomas Niederkrotenthaler[3], David Garcia[1,3,4]

**Affiliations:**

1. Section for Science of Complex Systems, Center for Medical Statistics, Informatics and Intelligent Systems, Medical University of Vienna, Austria
2. Unit Suicide Research & Mental Health Promotion, Department of Social and Preventive Medicine, Center for Public Health, Medical University of Vienna, Austria
3. Complexity Science Hub Vienna, Austria
4. Institute of Interactive Systems and Data Science, Graz University of Technology, Graz, Austria
5. Institute of Globally Distributed Open Research and Education, Austria
6. Institute of Information Systems Engineering, Vienna University of Technology, Vienna, Austria

## Dataset for Training Machine Learning Models

Time period: Because we initially started annotating tweets directly on the Crimson Hexagon platform, before downloading tweets for the target period (2013-2020), the training set includes a small fraction of 72 tweets from the years 2009 to 2012.

## List of search terms used to detect tweets for the training dataset

We started exploring the full dataset of tweets about suicide using keywords or phrases we suspected to be indicative of each of the five initial broad categories. After brainstorming some keywords (e.g. commit, lifeline, hope), we also used word frequency plots to identify typical terms in each category, and to search further example tweets with similar content. The list below includes all of these keywords, both those that worked well, as well as those that did not return many (too specific) or too many examples (too general). Asterisks indicate that all possible word endings were included, and terms in brackets indicate which examples we hoped to find with a keyword in cases where this is not obvious. We used no keywords for irrelevant tweets, given their high frequency.

**Suicide cases**

- Successful terms: commit*, found dead, news, heartbreaking, terrible, breaking news
- Too specific or general: police, jump*, hang*, hung, shot, shoot*, report*, suspect* (suspected suicide)

**Stories of coping**

- Successful terms: my story, my life, my experience, I (have) experienced, recover*, surviv*, strong, strength, thankful, okay, personal, I attempt*, my attempt, overwhelmed, fight, don’t want to die, grateful, despite, stay alive, suicidal thoughts
- Too specific or general: myself, thank, life, live, alive, try, story

**Prevention**

- Successful terms: lifeline, helpline suicide hotline, prevention, contact, help, support, protect, thoughts + struggling, alone, pain, anxiety, depression, warning signs, call, stay safe, stay alive, save + life (save someone’s life)
- Too specific or general: mental health, #mentalhealth, public health, first responders, #firstresponders, warning signs, call

**Awareness:**

- Successful terms: awareness, please retweet/copy, #mentalhealthawarenessmonth, #suicidepreventionweek, #suicideawareness, #nationalsuicidepreventionmonth,
- Too specific or general, or no notes on how successful they were: Risk, depressed, depression, anxiety, ptsd, loneliness, mental illness, trauma, homeless*, veterans, rape, sexual violence, message, talk, someone, feeling, upset, friends, reach out, stigma (#nostigma, #stopthestigma), struggles

## Text sequence length for model input

Figure S1 shows the distribution of tokens in the training set, based on which we determined the maximum sequence length for the models. The mean of the training tweet length is 25, the 95th percentile 57, and the 99th percentile 67 tokens. Therefore, we use a maximum sequence length of 80 tokens and cut off tokens beyond the maximum sequence length.

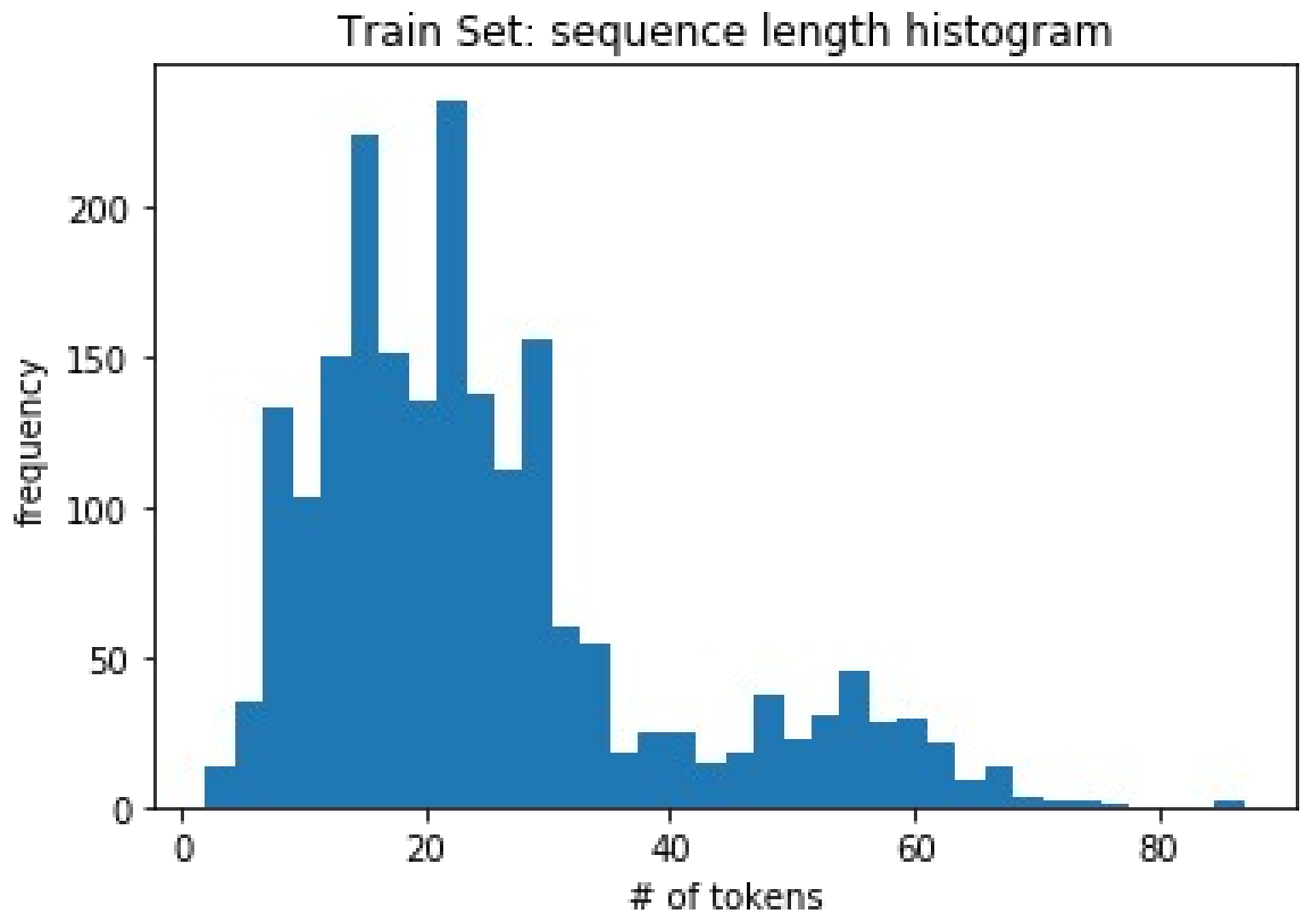

**Figure S1**. Histogram of the number of tokens (words etc) in the training set.

## Effect of different preprocessing strategies

For our main analyses, we only lowercase all words, and replaced URLs with http, and user mentions with @user. Additional NLP preprocessing best practises include removing digits, punctuation, stopwords, and lemmatization (grouping words according to their dictionary form). When working with longer texts, this eliminates frequent features that appear across most texts and provide little to no additional information. However, as tweets are already limited by a maximum content length of 280 characters, removing any information might worsen performance. We therefore explored how different preprocessing steps affect performance, first for lemmatization, and then for permutations of digit, stopword and punctuation removal. Lemmatization was not included in these permutations because punctuation and digits do not have lemmas, and the lemmas of stop words are often identical to the stop word (the, a , he, etc.).

We compare the different preprocessing strategies by training the base BERT model (12 layers, 12 attention heads, 110 million parameters, https://huggingface.co/bert-base-uncased) with a learning rate of 5e-5 for three epochs. Table S1 summarizes the results using the macro-averaged F1-score and average accuracy across all classes. Removing stopwords, and applying lemmatization, respectively, always decreased model performance. Removing only digits (strategy 3) performs similarly to the basic preprocessing (strategy 1), with a slightly higher validation F1 and slightly lower validation accuracy. Given that none of the additional preprocessing steps improved the performance, we apply strategy 1 for all analyses reported in the paper.

| | Preprocessing strategy | Validation set | | Test set | |
|---|---|---|---|---|---|
| | | *F1* | *Acc* | *F1* | *Acc* |
| 1 | BERT-base (lowercase, http, @user) | 0.72 | 0.76 | 0.72 | 0.73 |
| 2 | + lemmatization | 0.69 | 0.75 | 0.70 | 0.73 |
| 3 | - digits | 0.73 | 0.76 | 0.71 | 0.74 |
| 4 | - punctuation | 0.72 | 0.76 | 0.71 | 0.73 |
| 5 | - stopwords | 0.65 | 0.71 | 0.64 | 0.69 |
| 6 | - digits - punctuation | 0.73 | 0.76 | 0.70 | 0.72 |
| 7 | - digits - stopwords | 0.66 | 0.72 | 0.65 | 0.7 |
| 8 | - punctuation - stopword | 0.65 | 0.71 | 0.65 | 0.70 |
| 9 | - digits -stopwords - punctuation | 0.65 | 0.71 | 0.66 | 0.70 |

**Table S1.** Macro averaged F1 score and average accuracy with different preprocessing strategies, training with a learning rate=5e-5 and 3 epochs.

## BERT and XLNet performance in 5 separate runs

The performance of BERT and XLNet, with fixed seeds and parameters, varies slightly from run to run due to internal segmentation. Below, we report the results of five runs per task, which illustrates that variance was generally low.

**Task 1: Six main categories**

| | Validation set | | | | Test set | | | |
|---|---|---|---|---|---|---|---|---|
| **BERT** | **Pr** | **Re** | **F1** | **Acc** | **Pr** | **Re** | **F1** | **Acc** |
| Run 1 | 0.75 | 0.74 | 0.74 | 0.78 | 0.73 | 0.70 | 0.71 | 0.74 |
| Run 2 | 0.72 | 0.72 | 0.71 | 0.75 | 0.71 | 0.71 | 0.71 | 0.74 |
| Run 3 | 0.71 | 0.71 | 0.71 | 0.75 | 0.71 | 0.70 | 0.70 | 0.73 |
| Run 4 | 0.74 | 0.69 | 0.71 | 0.76 | 0.69 | 0.65 | 0.66 | 0.71 |
| Run 5 | 0.73 | 0.68 | 0.70 | 0.75 | 0.74 | 0.67 | 0.70 | 0.73 |
| **XLNet** | **Pr** | **Re** | **F1** | **Acc** | **Pr** | **Re** | **F1** | **Acc** |
| Run 1 | 0.75 | 0.74 | 0.75 | 0.78 | 0.71 | 0.72 | 0.71 | 0.75 |
| Run 2 | 0.71 | 0.72 | 0.71 | 0.76 | 0.69 | 0.72 | 0.71 | 0.71 |
| Run 3 | 0.76 | 0.74 | 0.75 | 0.78 | 0.72 | 0.70 | 0.71 | 0.76 |
| Run 4 | 0.76 | 0.71 | 0.72 | 0.77 | 0.73 | 0.69 | 0.70 | 0.76 |
| Run 5 | 0.72 | 0.74 | 0.73 | 0.76 | 0.71 | 0.74 | 0.72 | 0.72 |

**Table S2. Macro-averaged performance metrics and accuracy across all 6 categories on the validation and test set.** Pr=Precision, Re=Recall, Acc=Accuracy. Bold metrics are the best result per column. *Numbers in brackets indicate the learning rate, number of epochs, and the seed.

| Test set | Suicidal | | | Coping | | | Awareness | | |
|---|---|---|---|---|---|---|---|---|---|
| **BERT** | **Pr** | **Re** | **F1** | **Pr** | **Re** | **F1** | **Pr** | **Re** | **F1** |
| Run 1 | 0.62 | 0.49 | 0.55 | 0.82 | 0.67 | 0.74 | 0.69 | 0.72 | 0.70 |
| Run 2 | 0.62 | 0.46 | 0.53 | 0.70 | 0.71 | 0.71 | 0.74 | 0.79 | 0.76 |
| Run 3 | 0.60 | 0.46 | 0.52 | 0.78 | 0.69 | 0.73 | 0.66 | 0.73 | 0.69 |
| Run 4 | 0.46 | 0.40 | 0.43 | 0.74 | 0.67 | 0.70 | 0.77 | 0.55 | 0.64 |
| Run 5 | 0.61 | 0.45 | 0.52 | 0.77 | 0.70 | 0.73 | 0.68 | 0.71 | 0.69 |
| **XLNet** | **Pr** | **Re** | **F1** | **Pr** | **Re** | **F1** | **Pr** | **Re** | **F1** |
| Run 1 | 0.49 | 0.70 | 0.58 | 0.69 | 0.69 | 0.69 | 0.71 | 0.72 | 0.71 |
| Run 2 | 0.57 | 0.60 | 0.58 | 0.67 | 0.79 | 0.73 | 0.65 | 0.72 | 0.69 |
| Run 3 | 0.66 | 0.44 | 0.53 | 0.72 | 0.69 | 0.71 | 0.70 | 0.81 | 0.75 |
| Run 4 | 0.69 | 0.39 | 0.49 | 0.74 | 0.74 | 0.74 | 0.72 | 0.69 | 0.71 |
| Run 5 | 0.59 | 0.56 | 0.58 | 0.74 | 0.81 | 0.77 | 0.67 | 0.78 | 0.72 |

| | Prevention | | | Suicide cases | | | Irrelevant | | |
|---|---|---|---|---|---|---|---|---|---|
| **BERT** | **Pr** | **Re** | **F1** | **Pr** | **Re** | **F1** | **Pr** | **Re** | **F1** |
| Run 1 | 0.84 | 0.87 | 0.85 | 0.75 | 0.77 | 0.76 | 0.64 | 0.71 | 0.68 |
| Run 2 | 0.83 | 0.87 | 0.85 | 0.78 | 0.78 | 0.78 | 0.62 | 0.65 | 0.64 |
| Run 3 | 0.82 | 0.88 | 0.85 | 0.75 | 0.74 | 0.75 | 0.61 | 0.68 | 0.65 |
| Run 4 | 0.75 | 0.92 | 0.83 | 0.73 | 0.81 | 0.77 | 0.69 | 0.54 | 0.61 |
| Run 5 | 0.81 | 0.89 | 0.85 | 0.74 | 0.75 | 0.75 | 0.62 | 0.67 | 0.65 |
| **XLNet** | **Pr** | **Re** | **F1** | **Pr** | **Re** | **F1** | **Pr** | **Re** | **F1** |
| Run 1 | 0.85 | 0.85 | 0.85 | 0.79 | 0.74 | 0.76 | 0.71 | 0.63 | 0.67 |
| Run 2 | 0.83 | 0.87 | 0.85 | 0.77 | 0.71 | 0.74 | 0.65 | 0.65 | 0.65 |
| Run 3 | 0.81 | 0.88 | 0.84 | 0.78 | 0.78 | 0.78 | 0.67 | 0.63 | 0.65 |
| Run 4 | 0.81 | 0.86 | 0.83 | 0.74 | 0.82 | 0.78 | 0.71 | 0.62 | 0.66 |
| Run 5 | 0.81 | 0.90 | 0.85 | 0.80 | 0.71 | 0.75 | 0.65 | 0.68 | 0.67 |

**Table S3. Intra-class performance metrics for deep learning models on the test set.** Pr=Precision, Re=Recall, Acc=Accuracy.

**Task 2: About actual suicide**

| | Validation set | | | | Test set | | | |
|---|---|---|---|---|---|---|---|---|
| **BERT** | **Pr** | **Re** | **F1** | **Acc** | **Pr** | **Re** | **F1** | **Acc** |
| Run 1 | **0.86** | 0.80 | 0.82 | **0.88** | **0.85** | 0.81 | 0.83 | 0.88 |
| Run 2 | 0.89 | 0.80 | 0.83 | 0.89 | 0.87 | 0.78 | 0.81 | 0.88 |
| Run 3 | 0.83 | 0.81 | 0.82 | 0.87 | 0.87 | 0.83 | 0.85 | 0.89 |
| Run 4 | 0.83 | 0.81 | 0.82 | 0.87 | 0.85 | 0.83 | 0.84 | 0.89 |
| Run 5 | 0.84 | 0.86 | 0.85 | 0.88 | 0.81 | 0.82 | 0.82 | 0.86 |
| **XLNet** | **Pr** | **Re** | **F1** | **Acc** | **Pr** | **Re** | **F1** | **Acc** |
| Run 1 | 0.84 | 0.79 | 0.81 | 0.87 | 0.84 | 0.81 | 0.83 | 0.88 |
| Run 2 | 0.84 | 0.78 | 0.80 | 0.87 | 0.85 | 0.79 | 0.81 | 0.88 |
| Run 3 | 0.85 | 0.78 | 0.80 | 0.87 | 0.83 | 0.78 | 0.80 | 0.87 |
| Run 4 | 0.84 | 0.78 | 0.80 | 0.87 | 0.82 | 0.81 | 0.81 | 0.86 |
| Run 5 | 0.83 | 0.79 | 0.81 | 0.87 | 0.82 | 0.82 | 0.82 | 0.87 |

**Table S4. Macro-averaged performance metrics and accuracy for Task 2 (about suicide vs. off-topic) on the validation and test set.** Pr=Precision, Re=Recall, Acc=Accuracy.

| Test set | About suicide | | | Off-topic | | |
|---|---|---|---|---|---|---|
| **BERT** | **Pr** | **Re** | **F1** | **Pr** | **Re** | **F1** |
| Run 1 | 0.90 | 0.95 | 0.92 | 0.80 | 0.67 | 0.73 |
| Run 2 | 0.88 | 0.97 | 0.92 | 0.85 | 0.59 | 0.70 |
| Run 3 | 0.92 | 0.95 | 0.93 | 0.82 | 0.72 | 0.76 |
| Run 4 | 0.91 | 0.94 | 0.93 | 0.80 | 0.71 | 0.75 |
| Run 5 | 0.91 | 0.90 | 0.91 | 0.72 | 0.73 | 0.73 |
| **XLNet** | **Pr** | **Re** | **F1** | **Pr** | **Re** | **F1** |
| Run 1 | 0.91 | 0.94 | 0.92 | 0.78 | 0.69 | 0.73 |
| Run 2 | 0.89 | 0.96 | 0.92 | 0.81 | 0.62 | 0.70 |
| Run 3 | 0.89 | 0.95 | 0.92 | 0.78 | 0.61 | 0.69 |
| Run 4 | 0.91 | 0.91 | 0.91 | 0.72 | 0.72 | 0.72 |
| Run 5 | 0.91 | 0.91 | 0.91 | 0.72 | 0.72 | 0.72 |

**Table S5. Intra-class performance metrics for Task 2 (about suicide vs. off-topic) on the test set.** Pr=Precision, Re=Recall, Acc=Accuracy.

## Model vs. human performance

| Metric | Human-model | Suicidal | Coping | Awareness | Prevention | Suicide cases | Irrelevant |
|---|---|---|---|---|---|---|---|
| Pr | Coder 1 - BERT | 0.60 | 0.74 | 0.48 | 0.78 | 0.87 | 0.53 |
| | Coder 2 - BERT | 0.64 | 0.68 | 0.55 | 0.86 | 0.86 | 0.46 |
| | Coder 1 - Coder 2 | 0.82 | 0.82 | 0.71 | 0.90 | 0.82 | 0.52 |
| Re | Coder 1 - BERT | 0.67 | 0.76 | 0.87 | 0.82 | 0.78 | 0.29 |
| | Coder 2 - BERT | 0.58 | 0.76 | 0.91 | 0.76 | 0.82 | 0.35 |
| | Coder 1 - Coder 2 | 0.67 | 0.88 | 0.64 | 0.75 | 0.87 | 0.70 |
| F1 | Coder 1 - BERT | 0.63 | 0.75 | 0.62 | 0.80 | 0.82 | 0.38 |
| | Coder 2 - BERT | 0.61 | 0.72 | 0.69 | 0.81 | 0.84 | 0.40 |
| | Coder 1 - Coder 2 | 0.74 | 0.85 | 0.67 | 0.82 | 0.85 | 0.60 |

**Table S6. Inter-rater reliability per tweet category between two human coders and the BERT model, and between human coders in Task 2 in the reliability dataset.** Pr=precision. Re=recall. The human coder listed before the hyphen in the column human-model was used as the ground truth. For each metric, the first two lines compare the model versus each coder, with the coder as the ground truth. The third line reports the same metrics for coder 2 compared to coder 1 as the ground truth.

### Word clouds for all tweets in categories about actual suicide

All below word clouds are based on tweets in the entire labelled dataset (including the train- and test set).

**Suicidal ideation and behavior**

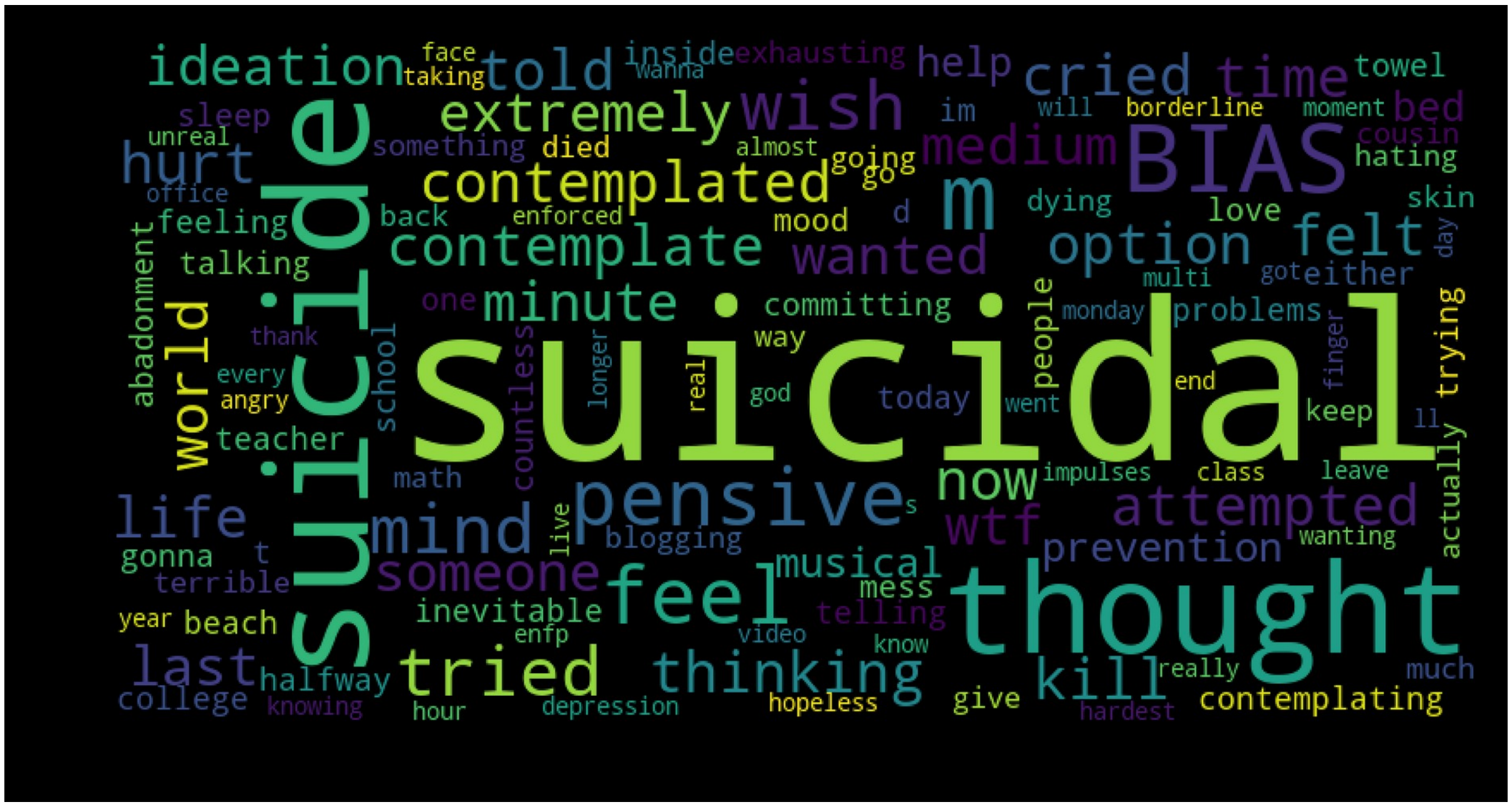

**Coping**

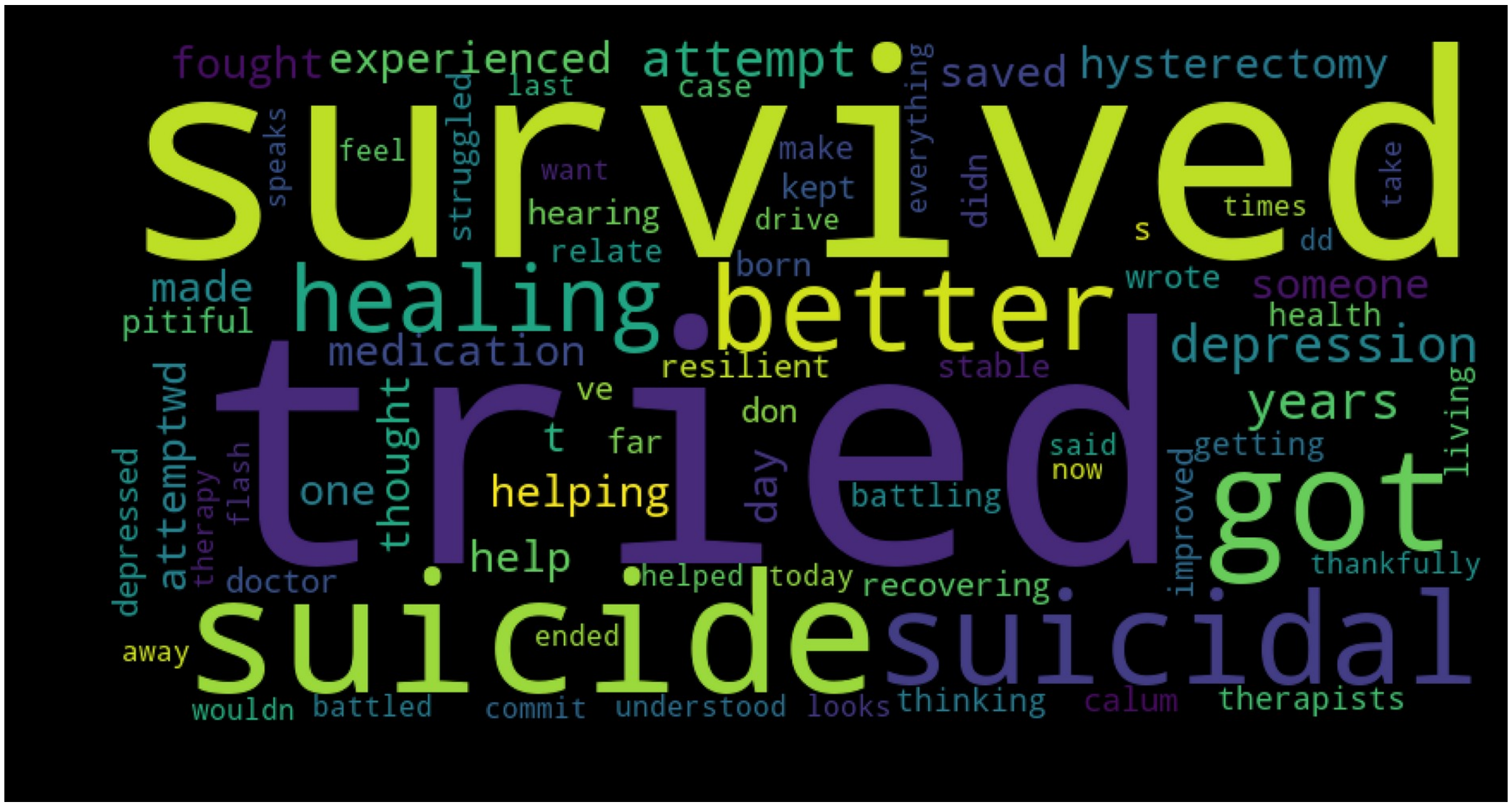

**Prevention**

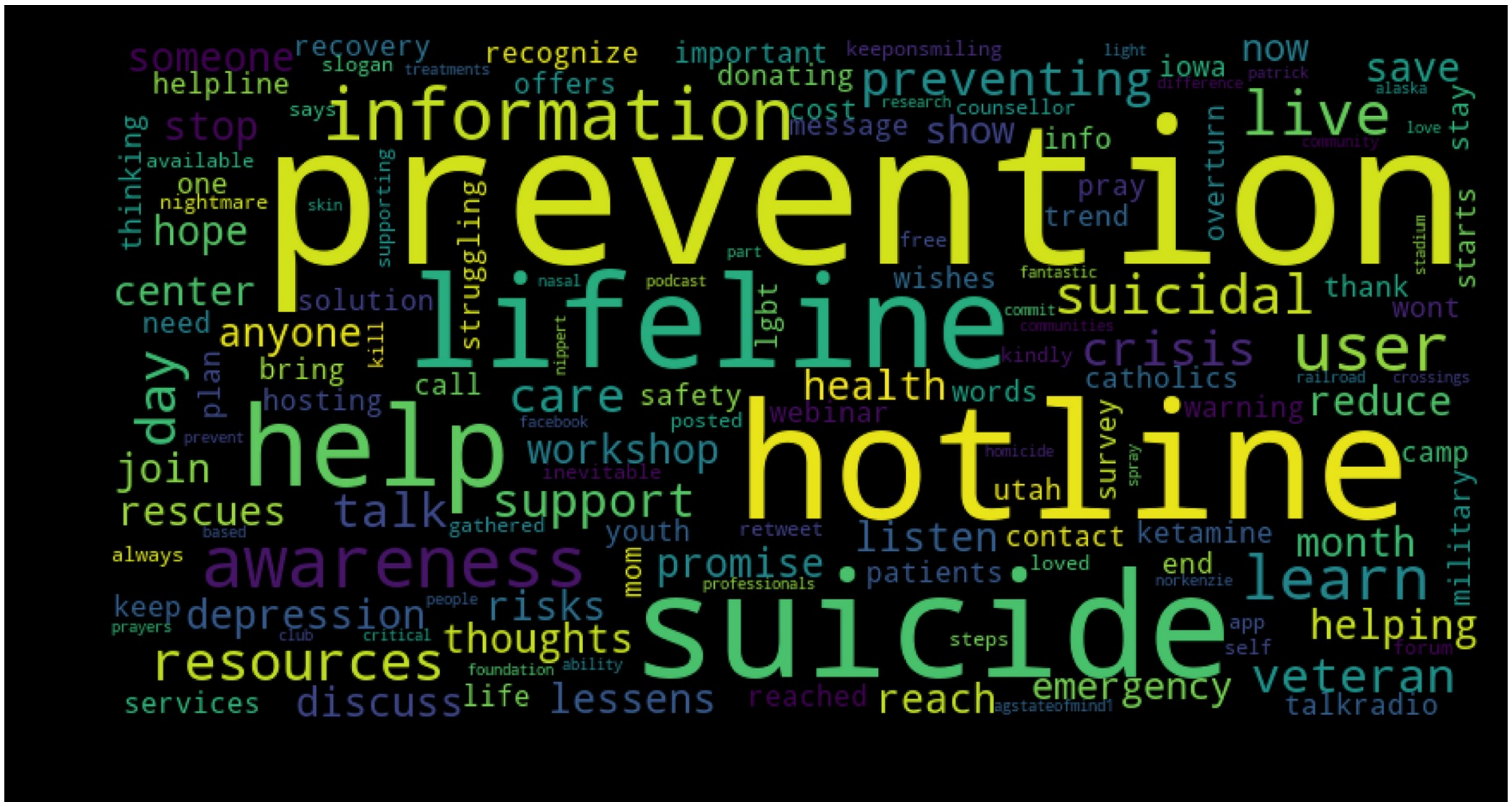

**Awareness**

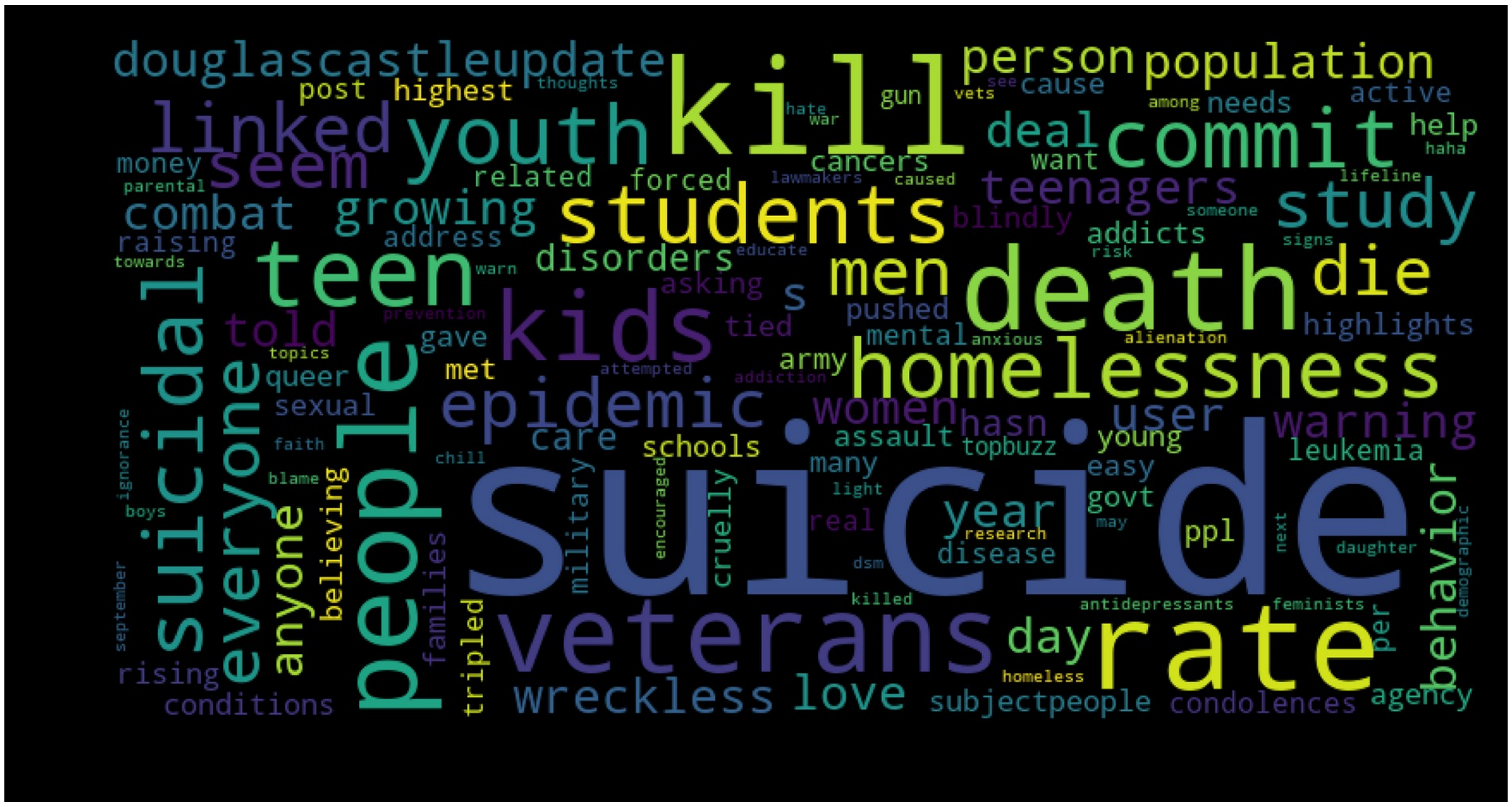

**Suicide cases**

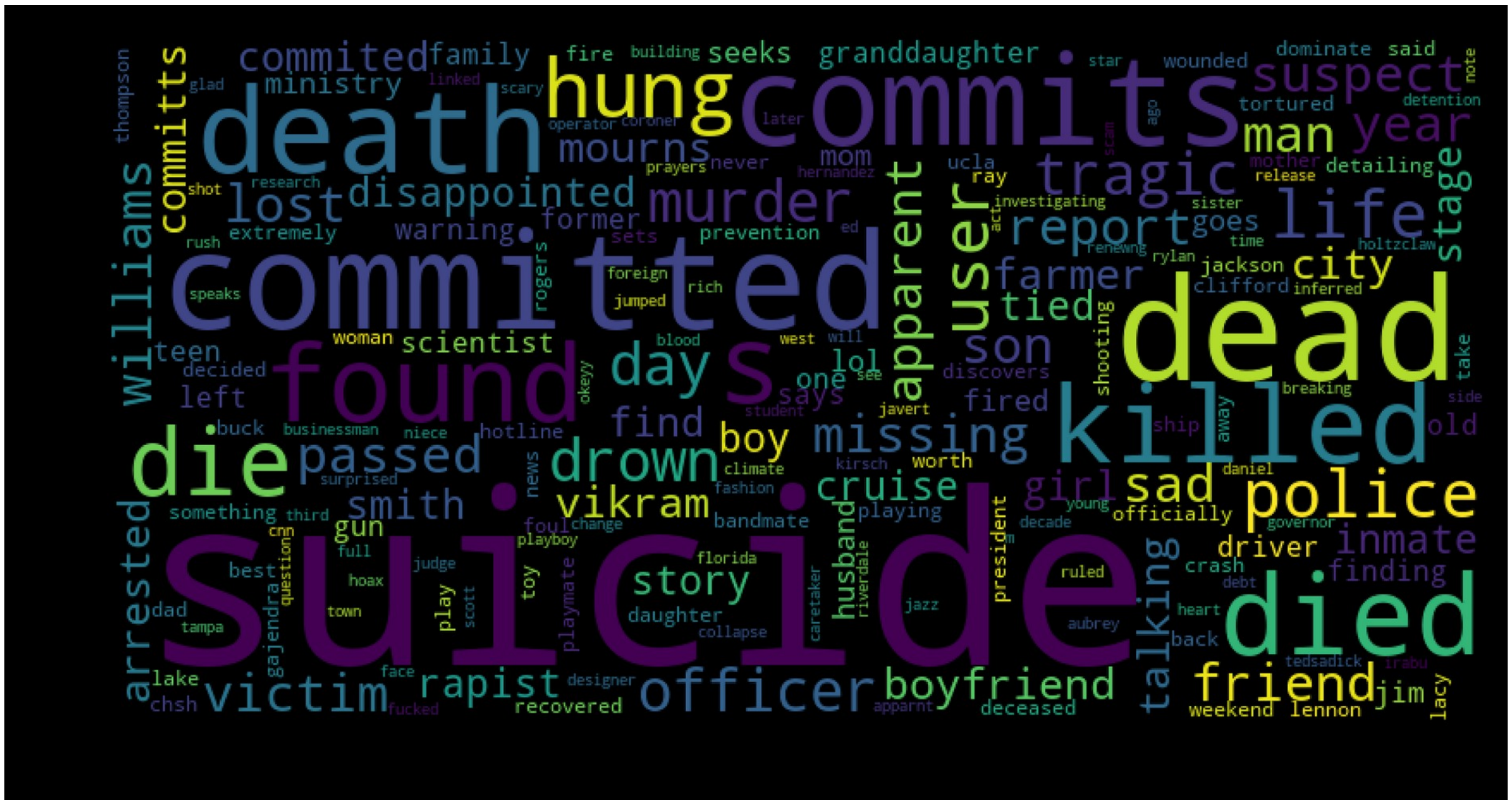

**Suicide other**

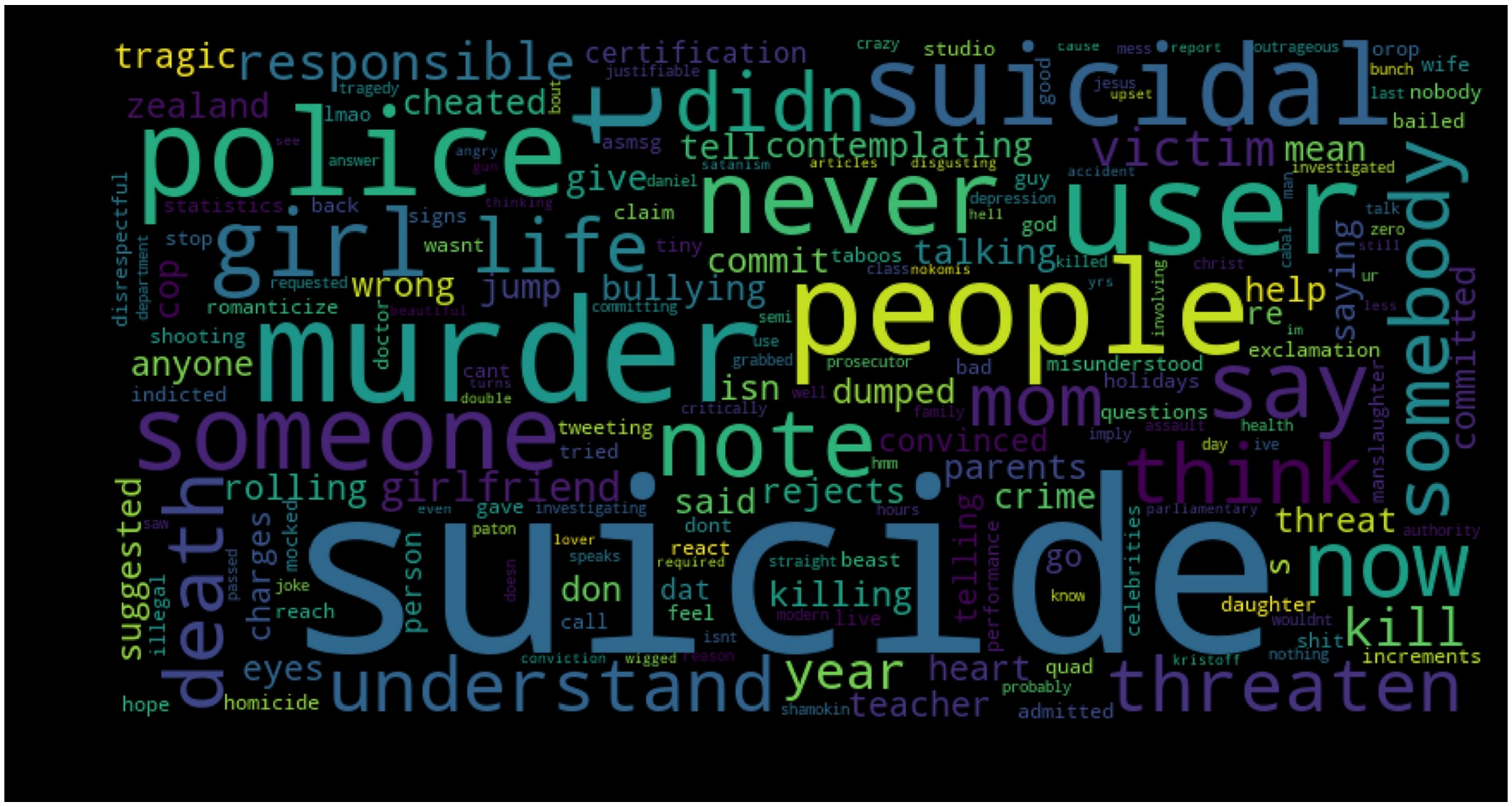

**Bereaved negative**

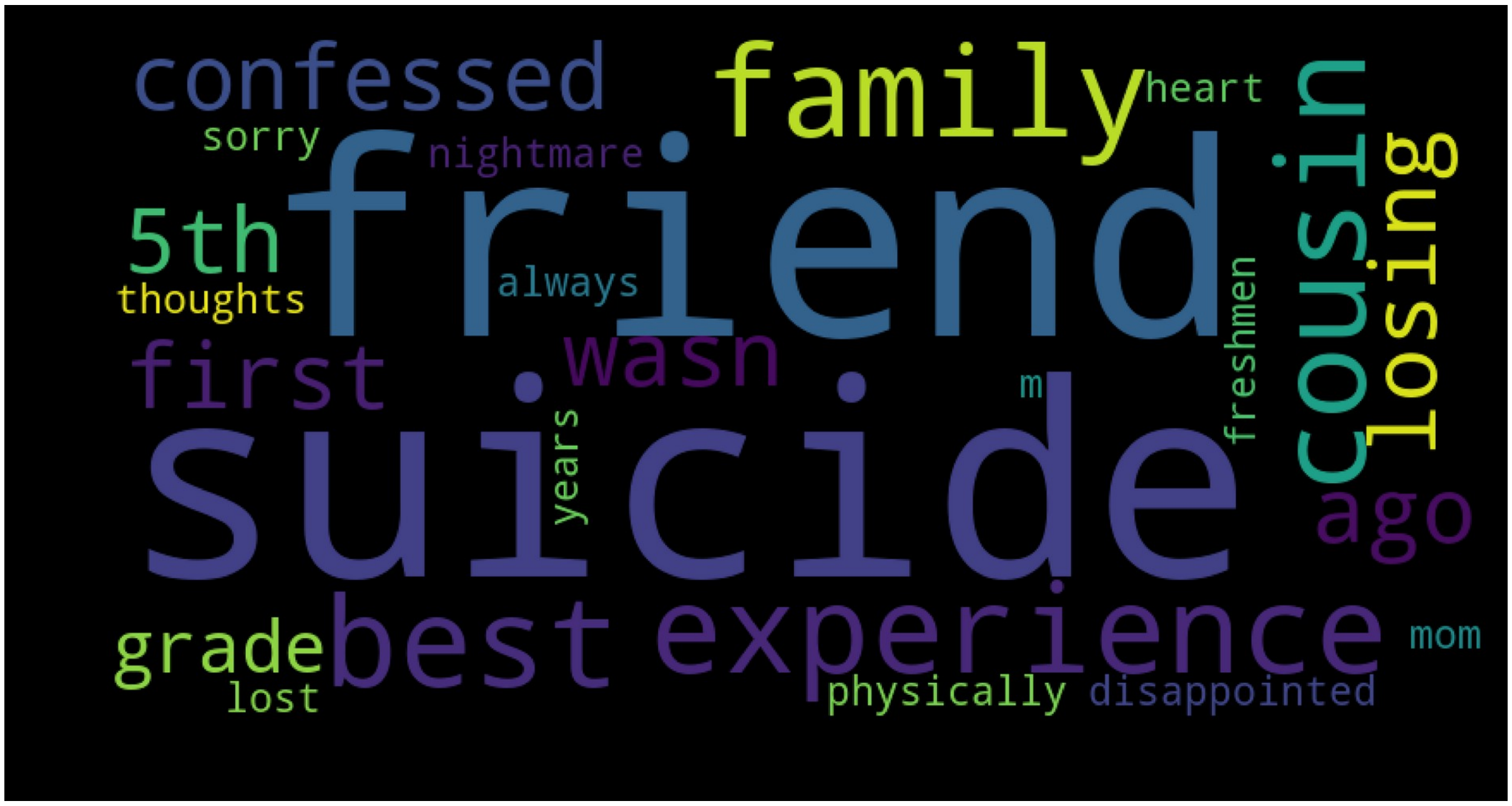

**Bereaved coping**

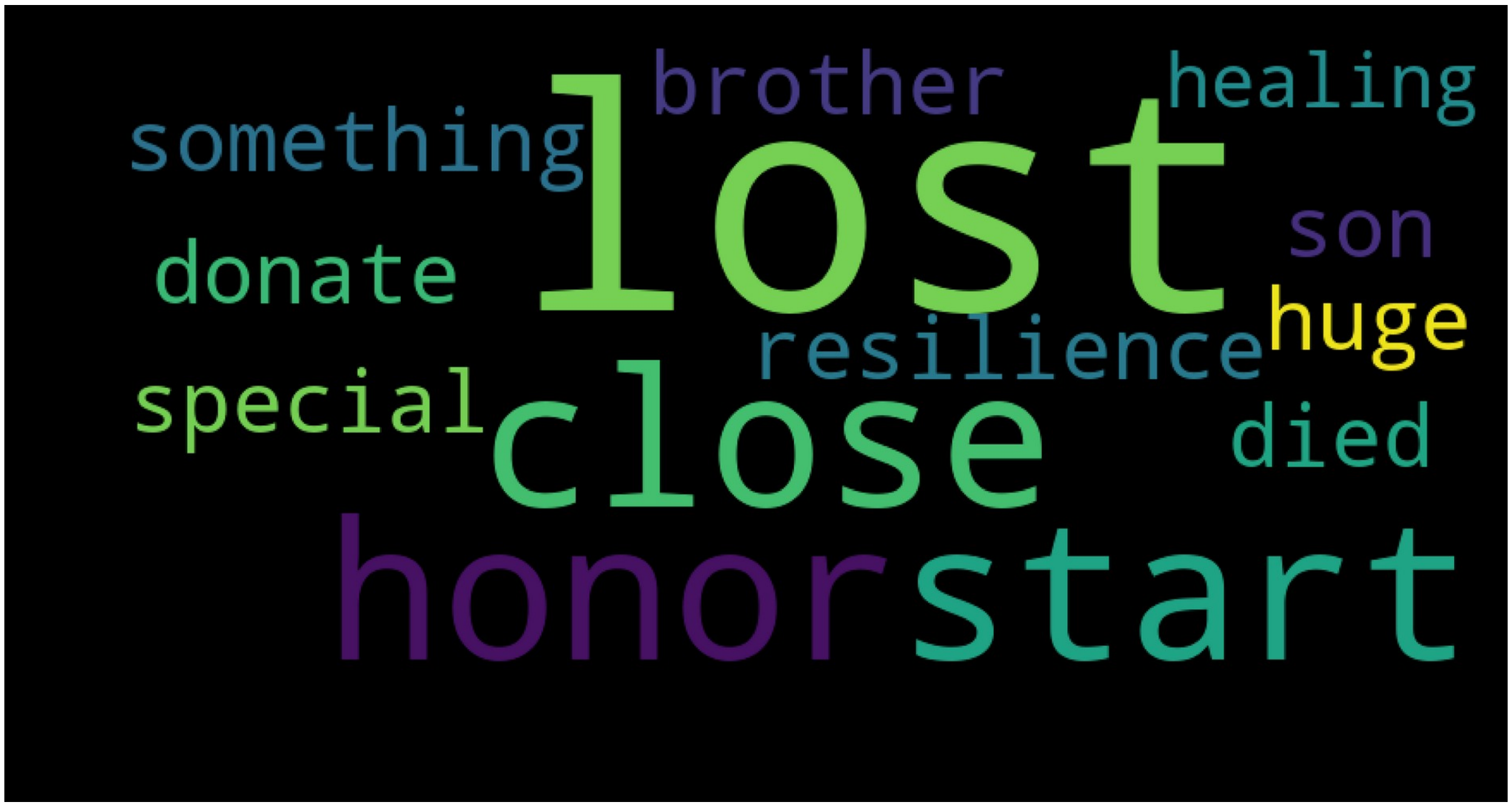

**News suicidal ideation and attempts**

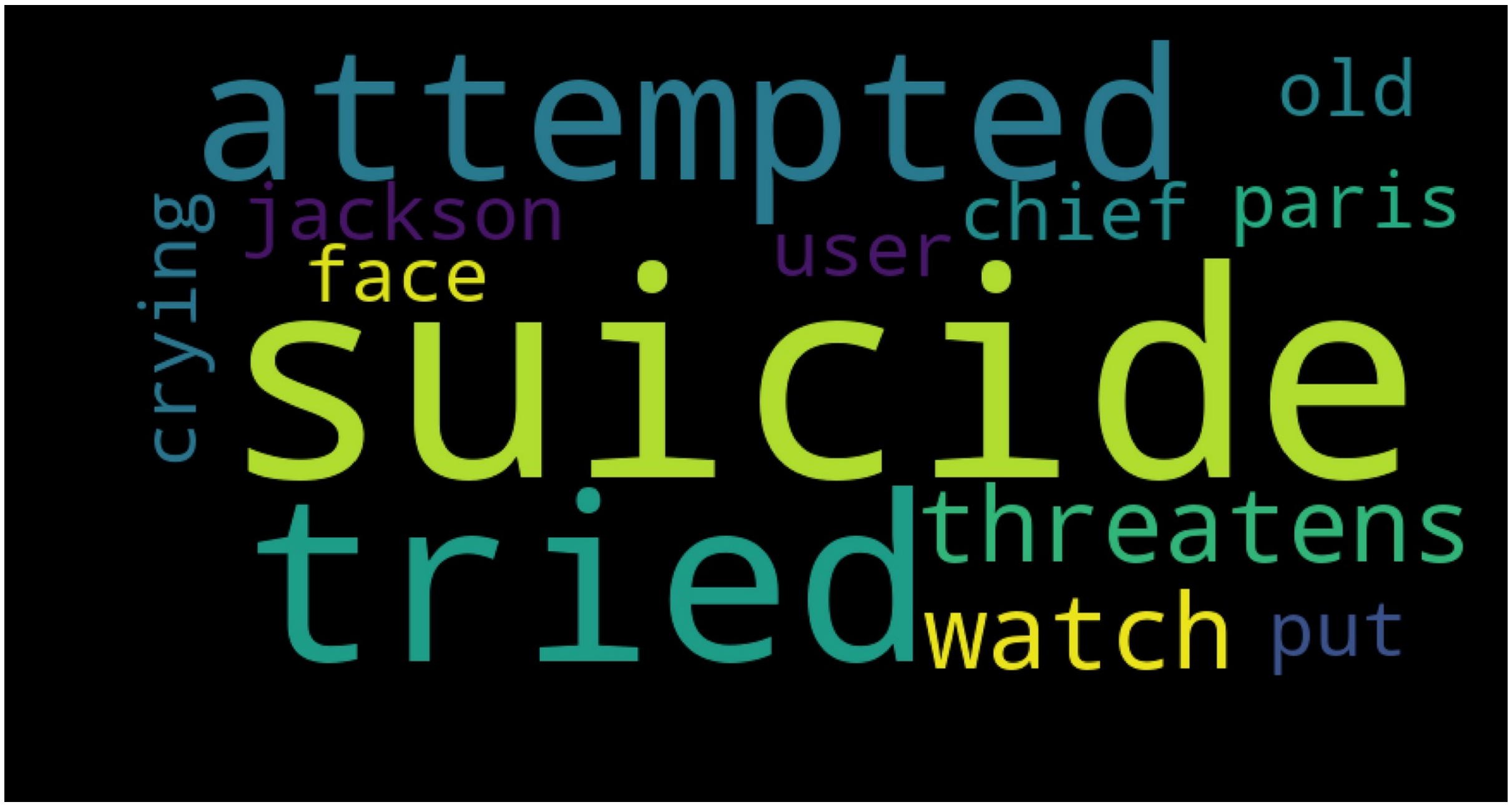

**News coping**

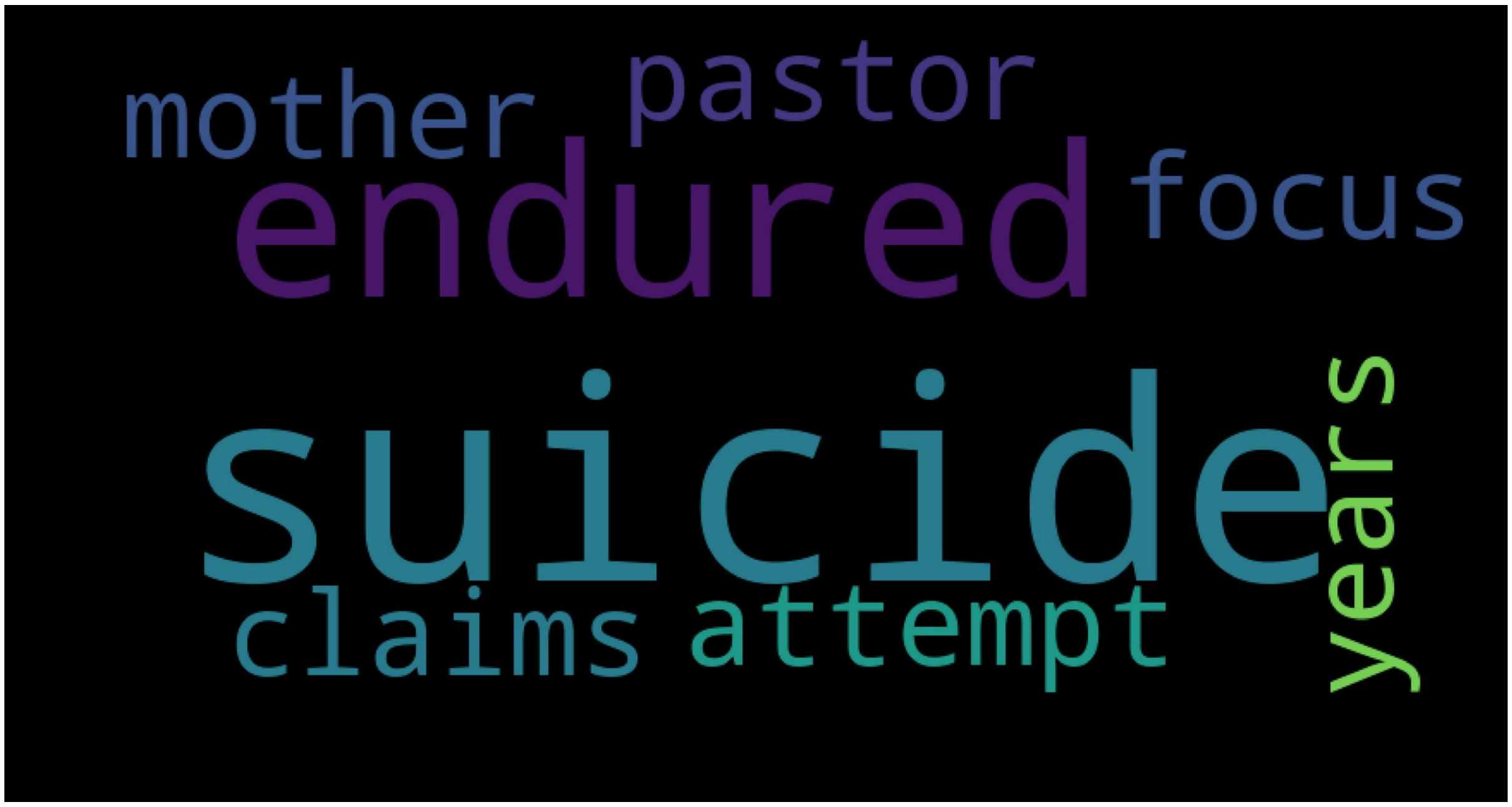

# Appendix 2

# Annotation scheme with detailed category definitions and tweet examples

**Authors**: Hannah Metzler[1,2,3,4,5], Hubert Baginski[3,6], Thomas Niederkrotenthaler[3], David Garcia[1,3,4]

**Affiliations:**

1. Section for Science of Complex Systems, Center for Medical Statistics, Informatics and Intelligent Systems, Medical University of Vienna, Austria
2. Unit Suicide Research & Mental Health Promotion, Department of Social and Preventive Medicine, Center for Public Health, Medical University of Vienna, Austria
3. Complexity Science Hub Vienna, Austria
4. Institute of Interactive Systems and Data Science, Graz University of Technology, Graz, Austria
5. Institute of Globally Distributed Open Research and Education, Austria
6. Institute of Information Systems Engineering, Vienna University of Technology, Vienna, Austria

# Overview of dimensions and categories

Tweet categories are organized along two dimensions, the type of message and the underlying perspective with regard to solutions and coping. The combination of these two dimensions divides postings into 10 categories of interest, and 2 irrelevant categories. A third dimension applies only to irrelevant tweets, and indicates if a tweet is *serious or not serious*. Dimensions and categories are described below, followed by detailed instructions and examples for tweet annotation.

| | | Underlying perspective | |
|---|---|---|---|
| **Message type** | | **Problem & Suffering** | **Solution & Coping** |
| **Personal experiences**<br>1st or 3rd person | | Suicidal ideation & attempts | Coping (Papageno) |
| **News about experiences & behavior** | | News suicidal ideation & attempts | News coping |
| **Experience of bereaved** | | Bereaved negative | Bereaved coping |
| **Case reports** | | Suicide cases (Werther) | Lives saved |
| **Calls for action** | | Awareness | Prevention |
| **Irrelevant** | **Suicide other** | Murder-suicides, history, fiction, not being suicidal, opinions... | |
| | **Off-topic** | Bombings, euthanasia, jokes, metaphors, band/song names... | |

## Dimensions

### Dimension 1: Message Type

We distinguished between six types of messages:

1. Personal stories about experiences: Experience of an affected individual either in first or third person perspective
2. News reports about suicidal experiences and behavior except cases, often about celebrities
3. Personal experiences of bereaved individuals: Describes the experience of a person who lost someone to suicide in first or third person perspective, including news reports.
4. Reports of particular completed or prevented suicide cases, often news reports
5. Calls for action: These are general statements calling for actions addressing the problem of suicide, and intending to spread problem awareness or prevention-related information.
6. Irrelevant: Message does not fit into any of the above categories. Here we distinguished between messages that were about suicide but did not fit in the above categories (suicide other), and those that were entirely off-topic (i.e. not about someone taking their life).

### Dimension 2: Underlying Perspective

Perspective refers to whether the tweet seems to communicate that there are hope and solutions for dealing with, and/ or potentially preventing, suicide. This may be expressed implicitly or explicitly. We distinguished:

1. Messages that frame suicide only as a problem and/or from an exclusively negative/suffering perspective.
2. Messages implying that solutions and ways of dealing with the problem exist, in a neutral or positive tone, but not negative. Any hint at a solution, alternative strategies, an attempt of coping, ways of becoming active or supporting efforts to fight suicide count.
3. Messages that do neither (only possible for messages in the irrelevant category)

Tweets in the category *suicide other* could be labeled as either of those as well as neither, and off-topic tweets were all labeled as neither, given that they focus on another topic.

**Dimension 3: Serious vs. Not Serious**

Only tweets that were clearly about suicide in the sense of someone's life ending counted as serious, whereas metaphors, exaggerations, sarcasm and jokes counted as not serious. If it was unclear whether a tweet was serious (e.g., sarcastic tweets, exaggerations), it was labeled as not serious. Only tweets in the two irrelevant categories could be labeled as not serious, given that all others are about actual suicide.

## Categories

- ***Suicidal ideation & attempts:*** Personal stories about an individuals negative experience with suicidal thoughts, related suffering (e.g. depression), suicidal communication and announcements, or suicide attempts, from the perspective of an affected individual, in 1st or 3rd person perspective.
- ***Coping***: Personal stories about an individuals experience with suicidal thoughts or a suicide attempt, with a sense of hope, recovery, coping, or mentioning an alternative to suicide. The sentiment does not have to be positive. A neutral tone, or talking about difficult experiences with a sense of coping or mentioning recovery, is sufficient.Previous research suggests such messages may have a *Papageno effect.*
- ***News suicidal ideation & attempts:*** News reports about suicidal experiences without any mention of coping, including reports on suicidal ideation, suicide attempts, announcements of suicide, someone being put on "suicide watch", etc., often about celebrities
- **News coping:** News reports about attempted or successful coping with or recovering from a suicidal crises, often about celebrities
- ***Bereaved negative:*** Describes the suffering or purely negative experience of a person who lost someone to suicide, including depression, grief, loss etc. These tweets necessarily refer to a suicide case, but are labeled as bereaved as long as they focus on the experience of bereaved individuals.
- ***Bereaved coping:*** Describes the experience of a bereaved person with a sense of hope, recovery or coping. The sentiment does not have to be positive. A neutral tone, or talking about difficult experiences with a sense of coping or mentioning recovery, is sufficient.
- ***Suicide cases:*** About an individual suicide, or a timely or geographical suicide cluster. Suicide cases have priority over definition criteria of other categories (except tweets focusing on bereaved individuals which are always related to a suicide case). Previous research suggests such messages on individual suicide deaths (especially about celebrities) may have a *Werther effect*.
- ***Lives saved:*** News report or personal message about someone saving a life. In contrast to prevention tweets, these lives are often being saved coincidentally.
- ***Awareness:*** Tweets intending to spread awareness for the problem of suicide, often focusing on high suicide rates or associations with bullying, racism, depression, veterans etc. without hinting at any solution. These are often reports of research findings or suicide statistics. Mere expressions of intent to help count as awareness, but any hint at something that can be done counts as prevention.
- ***Prevention:*** Tweets spreading information about a solution or an attempt to solve the problem of suicide, including prevention at an individual (e.g. do not leave people alone in crisis situations) or public health level (e.g. safety nets on bridges). Hinting at a solution or a way of dealing with the problem is enough. No specific action needs to be described. These tweets often include a helpline number. Announcements of prevention events and broad recommendations for actions also count: donations, prayers with a focus on a solution for suicide, being there for someone, telling people that they matter, taking a course about suicide prevention, warning signs to watch out for,
- ***Suicide other:*** Anything about suicide but not clearly related to any other above category, including murder-suicides, confident statements that something was not a suicide, convincing statements of not being suicidal, historical tweets about suicides that are a minimum of 40 ago (e.g. about Hitler's suicide), movies, books, novels, fiction about suicide

- ***Off-topic:*** all messages that use the term suicide in a context other than someone taking their own life due to suicidal thoughts. This includes messages on euthanasia, suicide bombing, and suicide attacks, messages that are (suspected) jokes, irony, sarcasm, flippant remarks, or really unclear in terms of authenticity, messages that use suicidal/suicide to exaggerate an emotional experience (unclear if serious), or as a methaphor (e.g. political/financial/career suicide, suicide workout, suicidal immigration policies...), and messages about "suicidal animals" (e.g. killed by car).

For cases of tweets that cannot be clearly assigned to only one category, the instructions below include prioritization rules. These are based on:

- Strength evidence for this content type being associated with suicide rates from previous research (e.g., We have stronger evidence that suicide case reports may have harmful effects on suicide rates compared to awareness tweets, so if an awareness tweets includes a case report, this is labeled as suicide case.)
- Logical necessity (e.g., experiences of bereaved necessarily include a suicide case, but are only labeled as suicide case if they focus on the case, rather than the experience of the bereaved individual)
- The underlying perspective of the tweet: coping/solution oriented tweets might still mention a problem and suffering, since coping with suicide includes dealing with negative experiences. We therefore generally prioritize the coping/solution oriented category over the problem-centered category within the same message type when both elements are present (e.g., within personal stories on suicide, coping stories were prioritized over suicidal ideation stories).

# Detailed instructions for annotating tweets

The below tweet examples, if not published by news agencies, have been anonymised by replacing some words with synonyms, changing punctuation and deleting user names and links, so that the user's who published them on Twitter cannot be identified.

## 1) Personal experiences of affected individuals

- Experience of an affected individual either in first or third person perspective, indicated by Coping1/Suicidality1 vs. Coping3/Suicidality3
- Whenever a personal experience is mentioned, this has priority over awareness- or prevention-related statements in the same tweet. We code as personal or bereaved experience.
- If 1st and 3rd person perspectives are mixed, in a tweet from a third person perspective that quotes a 1st person perspective, code as 1st person.
- These personal experience categories do not apply in the following cases:
    - General statements on suicidal thoughts do not qualify
    - Personal experience of bereaved individuals go in a separate category
    - Any thought that is really unclear in terms of authenticity should go under off-topic (sarcastic, using suicidal to exaggerate an emotional experience)
    - If they seem posted by news outlets, they go into category 5.2: Coping news

### 1.1) Suicidal stories: Stories of suicidal ideation and attempts

- About individual suicidal thoughts or suicide attempt from the perspective of an affected individual, negative experiences and thoughts, suffering, depression, …
- Suicidal communication/announcements of suicide

*Examples*

- Suicidality1: It really fucking pissed me off because I am depressed and suicidal, but she of course doesn't know that.
- Suicidality1: My mother thinks that I am suicidal. She's not wrong.
- Suicidality1: Suicidal Thoughts.I will pretend things are okay as I always do.
- Suicidality3: My best friend just left the hospital after another suicide attempt. I knew something bad had happened...
- Example for exaggerations, where it is not clear if the writer is actually suicidal or making a joke/using suicide as a metaphor for feeling really bad:
    - The way I feel I want to commit suicide right fucking now!
    - My suicide note will be a math worksheet. I can't do it anymore
    - These people make me go suicidal.

#### 1.2) Coping stories: Stories of hope, recovery, and coping

- About an individuals experience with suicidal thoughts or a suicide attempt, with a sense of hope, recovery, coping, or mentionning an alternative to suicide.
- Does not have to be positive, but cannot be negative, neutral is enough. If not clearly negative, priority is on coping. These tweets could have a Papageno effect.
- Tweets that clearly indicate the experience of suicidal ideation or behavior lies in the past also belong here, because they suggest the person survived.

*Examples*

- Coping1: I realized that I did not want to die. I just did not want to continue my life as I knew it.
- Coping1: I simply don't feel valued. Sometimes, the ONLY thing that kept me from suicide was my dad and my love for myself that i had to find.
- Coping3: I have talked someone out of suicide, it is not that fucking scary.
- Coping 1 - expressing coping implicitly by talking about suicidal ideation or attempt in the past:
    - Back on November 5th I tried to commit suicide. I had no intent to continue living.
    - I was suicidal in the 6th and the 11th grade.

## 2) Experience of bereaved individuals

- Describes the experience of someone losing a close one to suicide from a first or third person perspective of bereaved individuals, including news reports about bereaved individuals.
- Simply mentioning someone’s suicide is not sufficient, the tweet needs to describe or implicitly hint at the bereaved person’s emotional experience related to the suicide.
- General statements (e.g., RIP followed by the name of a deceased person) don‘t qualify.
- These tweets necessarily refer to a suicide case, but are only labeled as suicide case rather than experience of bereaved when focusing on the suicide rather than the experience of bereavement.

#### 2.1) Bereaved negative: negative experience, suffering

- About a bereaved individuals negative experience/suffering with the suicide of a close one. Suffering, Depression, desperation, grief, loss, without a sense of coping or hope.
- If a tweet includes both coping/improvement and suffering, label it as coping.

*Examples*

- I am very sensitive about suicidal comments like "kill yourself". My best friend did this. I’ll kick your ass.

- The little brother of my best friend just attempted suicide. I’m about to start crying. Oh my god.
- Implicitly hints at suffering: 4 years ago my ex-girlfriend committed suicide. [Loudly crying face] God please make sure she is okay. :'( [folded hands]

**2.2) Bereaved coping: experience of coping, hope, recovery**

- About a bereaved individuals experience with the suicide of a close one, with a sense of hope, recovery, and coping, every attempt of coping counts (outcome need not be positive)
- Does not have to be positive, but cannot be negative, neutral is enough. If not exclusively negative, label it as coping.

*Examples*

- My girlfriend killed herself 12 years ago. Griefing myself, I watched others devastated by grief fall into a dark hole and never recover. So I chose to open my heart, deepen compassion and meditation, and go toward service. These were the gifts of grief. My life is better for all of it.
- The actor opened up about his mother's scary suicide attempt and how he quickly saved her from dying.
- Everyone has a story. After suicide of a friend, I wrote #OnPoint. Let's also write your story together!

Examples that do not qualify because they do not describe the bereaved person’s experience:

- My mom committed suicide 20 years ago. He had PTSD and was a Vietnam veteran, and this is really really important.
- My brother killed himself 7 years ago. I just found out I’m the beneficiary of his life insurance. What should I know?

## 3) Suicide cases

**3.1) Suicide cases (Werther)**

- About an individual suicide, or a suicide cluster (timely or geographical)
- These tweets could have a Werther effect.
- Suicide cases have priority over definition criteria of other categories:
  - If it is about prevention or awareness but still obviously related to a suicide or attempt, code here. Even if lifeline is mentioned or a solution focus is present, if a suicide case is tweeted, it belongs here. This means Werther tweets can very rarely have a solution focus, i.e. receive the label 2 in the column focus.
  - Only bereaved stories about a suicide with a clear focus on personal experience have priority over Werther. Personal stories about attempts belong into suicidality1/3.
  - If doubts are expressed if a death was a suicide , we still code it as a suicide case. E.g. „apparent suicide“, „possible suicide“, „suspected suicide“. (An individual at risk would perceive it as a suicide case.)
- Suicide by cop also qualifies
- Suicides of villains or criminals also qualifies

*Examples*:

- Staten Island boy commits suicide after bullying ignored @TwURL
- Married man attacks TN college girl, commits suicide: A married man who… @TwURL #SuryaRay #India
- It is a beautiful cliff. Many people commit suicide here.

- Villain: Steuben County man accused of murder dies by suicide.
- Suicide by cop: Utah police: Man killed by officers in 2014 intended to 'commit suicide by cop'
- Prevention-related but mentioning a suicide case, therfore code here: What a tremendous loss with the passing of Kate Spade. My condolences to her family and loved ones. Depression does not discriminate. National Suicide Prevention Lifeline: 1-800-273-8255
- Expressions of doubt:
  - Death of Lennon Lacy ruled a suicide by the FBI, questions still linger
  - Former Patriot Aaron Hernandez found dead in prison in apparent suicide

### 3.2) Life saved

- News reports about someone saving a life, opposite of Werther tweets
- In contrast to prevention tweets, these lives are often being saved incidentally.

*Examples*

- Authorities: Kayaker rescues man who jumped from bridge: Authorities say a man who attempted suicide by jumpin...
- All the commotion up by Callery Park is LPD trying to help a suicidal 16-year-old girl. She's safe now, but sounds like she's in bad shape
- thedailybeast: Dolphins saved a girl from committing suicide. Susan Casey explains how it happened here: …
- #MLB Wire: How a former Cowboys player helped stop John Daly from committing suicide

## 4) Calls for action

- Calls for action: These are general statements calling for actions addressing the problem of suicide, intending to spread awareness or prevention-related information.
- If a personal story is told (affected individual or bereaved), the tweet does NOT count as awareness or prevention, even if it includes other content typical for these categories.
- If the below rules are not sufficient to distinguish between awareness and prevention tweets, because the tweet contains both elements, judge what the focus of the message is (focused on problem/suffering vs. solution). See the examples for distinguishing these two categories below.

### 4.1) Awareness

- Tweets focused on the problem of suicide and high suicide rates, or associations with bullying, racism, depression, veterans etc. without hinting at any solution.
- These tweets might want to help prevent but give no recommendation on what to do and how.
- Often includes research findings that do not explicitely say what an individual can do to prevent suicide. These research findings do not need to be written with the intention to help – pure information/updates are awareness tweets too.
- Often these tweets ask other Twitter users to retweet information. Mere retweeting does not count as a solution, because it does not hint at anything that can be done to prevent suicide. Therefore, these are awareness rather than prevention tweets.

*Examples Awareness:*

- Suicide risk linked to insomnia, alcohol use, study shows: Insomnia symptoms mediate the relationship between ... @TwURL
- Kinda sad & creepy. Suicide forest, Japan. 200 people attempt suicide in this forest every year.

- EverydayHealth: ABC News Chief Medical Correspondent drjashton's new book shines a much-needed spotlight on the pain and struggles faced by people who have lost a loved one to death by suicide @TwURL
- Reblog if you would be devastated if you found out one of your followers committed suicide. - iiamylou:... @TwURL
- Research finding: Nearly 78 percent of the 45,000 people who kill themselves every year in the United States are men. Here’s what the research on masculinity tell us about this crisis: @TwURL
- 22 veterans commit suicide every day. Would at least 22 of my Twitter friends please copy this and retweet?

**4.2) Prevention**

- Tweet is on prevention on individual level (e.g. do not leave people alone in crisis situation) or public health level (e.g. safety nets on bridge)
- Focused on a solution or an attempt of solving the problem. Hinting at a solution or a way of dealing with the problem is enough. No specific action needs to be described.
- Tweets often have a help-line in focus. Mentioning the helpline is sufficient for indicating a first step towards a potential solution.
- General broad recommendation actions go here as well: donations, prayers with a focus on a solution for suicide, being there for someone, telling people that they matter, „call me“ without providing a phone number, taking a course/class/seminar about suicide prevention, warning signs to watch out for, announcing/advertising a prevention event

*Examples Prevention:*

- Individual prevention: When someone is suicidal you need to be there i don’t care what’s going on, drop what you are doing and be there for them.
- Individual and helpline: @SpringByrum RE: CluelessMomTho1. Please recommend her to call Suicide Prevention. It’s disturbing that people think it’s so taboo, they will not speak of it. The number is toll free and anonymous. God bless and have a blessed New Year.
- Helpline: You are loved, you are heard, you are not alone. Suicide Prevention 800-273-8255 Mental Health Hotline 888-991-4284 Center Against Sexual Assault 866-373-8300 National Alliance on Mental Illness 800-950-6264 National Sexual Assault 800-656-4673
- Helpchat: Chat - Idaho Suicide Prevention Hotline @TwURL
- Rather negative, but mentions helpline: #MensHealth is #MentalHealth. ‘74% of persons who died by suicide in Austin (2013-2017) were men - City of #Austin National Suicide Prevention Lifeline: 1-800-273-8255 City of Austin: Insured / Uninsured, visit @TwURL 24/7 Helpline 512-472-HELP (4357) @TwURL
- Unspecific, but solution-focused: 100 reasons not to commit suicide
- Event: #Seminar 07-17: Disarming the Suicidal Mind #BehavioralHealth #MO #StLouis @TwURL
- Public health level: Golden Gate Bridge suicide barrier funding approved - CBS News @TwURL
- Prevention news: Some Nebraska businesses are taking a creative approach to raise awareness around this issue, while also raising money to support local organizations that focus on veteran and service member suicide prevention. @TwURL

*Examples to distinguish Awareness vs. Prevention:*

- Solution focus that is totally irrelevant to suicide prevention, therefore awareness: Day 53 of 22 push-ups a day for infinite days. Spread the word that Veteran suicide is not the answer. #22PUSHUPS for #22KILL @TwURL
- Attempt at helping, no problem focus, therefore prevention: To everyone who has dealt with suicidal thoughts this past year: I'm glad you're still here :)

- Superficial solution, but not problem-oriented, therefore prevention: Suicide is a permanent solution to a temporary problem. Always remember that you are loved and NEVER alone.
- Prevention because more hope than problem focus: I just pray that everyone that has suicidal thoughts realizes that there is always a path to happiness. OR: SPECIAL PRAYER FOR THOSE WITH SUICIDAL THOUGHTS: Fear thou not
- Prevention, because there is a focus on listening, which is equivalent to offering a solution. Without the comment about listening, this would be awareness: Would three of my Facebook friends please copy and repost. I'm doing this to prove that someone is always listening. #SuicideAwareness
- Awareness, because it focuses on a problem (ambiguous case because it possibly hints that social inclusion on TV should be improved, but it does so by pointing out the problem, rather than saying "let's improve social inclusion"): @User 90% of Indigenous Australians deaths by suicide are under 30 year of age. Social inclusion on our TV matters.
- Awareness, because it only includes the word prevention, but does not hint at a solution: The 2nd leading cause of death in the US for ages 10-34 is suicide. Please reach out. #SuicidePrevention #theyellowelephant #oktobenotok #Mentalhealth @TwURL
- Awareness, because no solution, only retweeting (which does not prevent suicide): "@User: Every 17 seconds someone commits suicide because of cyber bullying. RT if you're against cyber bullying."

## 5) News about suicide (except suicide cases)

### 5.1) News about suicidal ideation and attempts

- News reports about suicidal ideation and attempts, often about celebrities, without any mention of coping
- Tweets (3rd person perspective) that mention „threat of suicide", or someone being put on suicide watch:

*Examples*

- "Threat of suicide" , or suicide watch:
- ENDANGERED EAGLE? Ex-rocker put on suicide watch after wife's death: @TwURL
- Vine Star Threatens Suicide After Video of Him Pressuring 16-Year Old Girl Into Oral Sex S… @TwURL
- Ray J's Girlfriend -- I Made a Mistake ... Suicide Threat Was Wrong: Ray J's girlfriend says she'd never kill … @TwURL
- Scarface @User speaks on suicidal thoughts @TwURL
- Iggy Azalea reveals she contemplated suicide @TwURL

### 5.2) News about coping and coping attempts

- News reports about coping with suicidal crises, recovery, attempts in the past etc., usually about celebrities
- Even if quotes from the first person perspective, code tweets as Coping_news rather than Coping1 because of their reach.
- Tweets about past suicidal experiences, because they suggest the person coped with the crisis.

*Examples*

- Gisele Bündchen Says She Battled Panic Attacks So Extreme She Considered Suicide: 'I Felt Powerless' Celebrities›Interviews The supermodel opens up about her battle with panic attacks. She's one of the highest-paid supermodels in the world, marrie… @TwURL
- Demi Lovato: I Was Suicidal at Age 7 @TwURL #knssradio

- Fox News Holly Madison's book reveals she contemplated suicide Fox News "Peepshow" star and former Playboy… @TwURL #Playboy

## 0) Irrelevant

### 0.1) Suicide other: about suicide but not in our categories of interest

- Anything about suicide but not clearly related to any other category goes here
- Murder-suicides
- Statements that something was not a suicide, with relatively high confidence.
- Someone convincingly saying they are not suicidal
- Personal opinions, e.g. people saying they don’t understand why people make jokes or laugh about suicide
- Historical tweets about suicides that are long ago (e.g. about Hitler), minimum 40 years (i.e. longer than Kurt Cobain, who might still have an influence)
- Movies, books, novels, fiction about suicide

*Examples*

- 12 Surprising Things That Happened On The Ground Immediately After Hitler Killed Himself @TwURL
- It's a response about anyone who has committed suicide, for any reason. Suicide is a personal choice, it's never the fault of anyone else. @TwURL
- Am I just thinking this is weird or does it seem like all teens are considering suicide?
- Fiction: I got more pussy after she killed herself than when she was my wife. This movie is fucked up.

### 0.2) Off-topic: not related to topic of actual suicide

- Jokes, suspected irony or sarcasm, flippant remark
- Using „ suicidal/suicide“ to exagerate an emotional experience (unclear if serious), suspected exaggeration
- Suicide and suicidal as methaphors (e.g. political/financial/career suicide, suicide workout, suicidal immigration policies...)
- Any thought that is really unclear in terms of authenticity goes here.
- Euthanasia, suicide bombing, suicide attacks, tweets are off-topic
- Tweets about “suicidal animals”, e.g. killed by car
- Animal suicides (see examples)

*Examples*

- You'll commit suicide trying to read my mind
- [HELP] last try until I take the suicide plunge to 10.5 on my iPhone 6 @TwURL
- Metaphors:
  - After 11 hours of meetings and about 3 hours in the car on suicidal roads, I realised it’s January.... now i'm depressed
  - "@User: Even hearing your name makes me want to commit suicide. I know the feeling!
  - Before you join the lemming line of b2b marketers flocking to all social: 15 Things to Consider Before You Commit Print Suicide @TwURL
  - He is commiting career suicide
  - Our country is committing policital suicide.

- Exaggeration:
  - After having the nail polish come off one of my nails I'm basically contemplating suicide
  - I forgot how suicidal math classes make me.
  - So I'm laying on my floor and I looked up to see this and almost committed suicide @TwURL
- Animal suicides:
  - A suicidal bird flew into my car.
  - I think my dog is suicidal.
  - Damn raccoon must have been suicidal, walking so slowly across the road, but I wasn't hitting something that big with my car!
  - At least three squirrels have tried to commit suicide via my car today